\documentclass[lettersize,journal]{IEEEtran}

\usepackage{bibunits}
\defaultbibliographystyle{IEEEtran}
\defaultbibliography{references}

\usepackage{amsmath,amssymb,amsfonts}
\usepackage{algorithmic}
\usepackage{graphicx}
\usepackage{textcomp}
\usepackage{xcolor}
\usepackage{subcaption}
\usepackage{float}
\usepackage{url}
\usepackage{bm}
\usepackage{booktabs}
\usepackage{hyperref}
\usepackage{svg}  
\usepackage{microtype}
\usepackage{amsthm}
\usepackage{mathtools}
\usepackage{tikz}
\usepackage{pgfplots}
\usepackage{pgfplotstable}
\usepgfplotslibrary{groupplots}
\usepackage[ruled,vlined,linesnumbered]{algorithm2e}
\usepackage{todonotes}

\usepackage{pdflscape}
\usepackage{array}
\usepackage{longtable}

\theoremstyle{definition}
\newtheorem{definition}{Definition}[section]

\def\BibTeX{{\rm B\kern-.05em{\sc i\kern-.025em b}\kern-.08em
    T\kern-.1667em\lower.7ex\hbox{E}\kern-.125emX}}
    
\begin{document}

\title{Investigating Scale Independent UCT\\ Exploration Factor Strategies}

\author{
  \IEEEauthorblockN{Robin Schmöcker - }
  \IEEEauthorblockA{\textit{Leibniz Universität Hannover - } \texttt{schmoecker@tnt.uni-hannover.de}}\\
  \and
  \IEEEauthorblockN{Christoph Schnell - }
  \IEEEauthorblockA{\textit{Leibniz Universität Hannover - } \texttt{schnell@tnt.uni-hannover.de}}\\
  \and
  \IEEEauthorblockN{Alexander Dockhorn - }
  \IEEEauthorblockA{\textit{University of Southern Denmark - } \texttt{ adoc@mmmi.sdu.dk}}\\
}


\maketitle


\begin{abstract}
  The Upper Confidence Bounds For Trees (UCT) algorithm is not agnostic to the reward scale of the game it is applied to. For zero-sum games with the sparse rewards of $\{-1,0,1\}$ at the end of the game, this is not a problem, but many games often feature dense rewards with hand-picked reward scales, causing a node's Q-value to span different magnitudes across different games. In this paper, we evaluate various strategies for adaptively choosing the UCT exploration constant $\lambda$, called $\lambda$-strategies, that are agnostic to the game's reward scale. These $\lambda$-strategies include those proposed in the literature as well as five new strategies. 
Given our experimental results, we recommend using one of our newly suggested $\lambda$-strategies, which is to choose $\lambda$ as $2 \cdot \sigma$ where $\sigma$ is the empirical standard deviation of all state-action pairs' Q-values of the search tree. This method outperforms existing $\lambda$-strategies across a wide range of tasks both in terms of a single parameter value and the peak performances obtained by optimizing all available parameters.
\end{abstract}


\section{Introduction}
\label{sec:intro}
Monte Carlo Tree Search (MCTS) \cite{BrownePWLCRTPSC12, Coulom06} is a popular model-based decision-making algorithm that is applicable to a wide range of games. A selection strategy within MCTS that is nowadays synonymously used with MCTS is Upper Confidence Bounds For Trees (UCT) \cite{KocsisS06}, which balances exploration and exploitation by employing the Upper Confidence Bounds (UCB) algorithm for multi-armed bandits as the tree policy \cite{KocsisS06}.

UCT, however, introduces one parameter, the exploration constant $\lambda > 0$ used for the UCB formula whose choice can significantly impact performance (see Section \ref{sec:exp}). While a single value for $\lambda$ may perform well across a variety of games with a similar scale for the Q-values, if $\lambda$ is not on the same scale as the node's Q-value, UCT may either degrade to uniformly picking actions during the tree policy (when $\lambda \gg |Q|$) or to the greedy policy (when $\lambda \ll |Q|$). This is problematic as the scale of the score which the agent tries to maximize is oftentimes arbitrarily chosen. The games that are later used for evaluation prove this point as they encompass scores ranging from one to four digits.

While several works discuss methods for choosing a $\lambda$ that adapts to the problem's scale \cite{OGAUCT,BonetG12,BallaF09, EyerichKH10, SchrittwieserAH20} these works, however, most treat this choice as an implementation detail. 

In this paper, we evaluate the performance of several of these strategies as well as propose new strategies for adaptively choosing $\lambda$ on a variety of single-player and two-player games and show that the methods suggested in the literature are outperformed by the methods introduced here. From here, we refer to any method for choosing $\lambda$ as a $\lambda$-strategy.
Since the goal of this paper is to recommend a drop-in $\lambda$-strategy replacement for Vanilla UCT (i.e., using a fixed $\lambda$ value), the
method had to fulfill four criteria, which are all satisfied by our newly proposed \textbf{Global Std} strategy that chooses $\lambda$ as $C \cdot \sigma$ where $C \in \mathbb{R}^+$ and $\sigma$ is the empirical standard deviation of all state-action pairs' Q-values.

 \noindent 1) Scale independence: If the reward function of a problem is multiplied by a constant $\mu > 0$, then this must not affect the policy induced by UCT.
 
 \noindent 2) The strategy must have little to no computational overhead.
 
 \noindent 3) There should be a single parameter of the $\lambda$-strategy that outperforms any single Vanilla UCT exploration constant choice across games. For Global Std, this is the case for $C=2$.
 
 \noindent 4) Optimizing over the parameters of the $\lambda$-strategy (if any parameters are introduced) on a per-task-level should outperform UCT when its exploration constant is also optimized. 
\\ \\
\noindent Our contributions can be summarized as follows.
\begin{itemize}
    \item We conducted a large-scale study on several $\lambda$-strategies from the literature as well as newly proposed ones by evaluating them on a plethora of single-player and two-player decision-making tasks with varying reward scales.
    \item One strategy that we proposed, namely Global Std, outperforms all methods from the literature both in terms of generalization and peak performance, whilst simultaneously satisfying the above-mentioned criteria.
    \item In particular, we show that Vanilla UCT, i.e., using a fixed exploration constant, performs poorly across the problems considered here that have vastly different reward scales.
\end{itemize}

The rest of this paper is structured as follows. Firstly, we present related work in \textbf{Section \ref{sec:related_work}}. Then, in \textbf{Section \ref{sec:method}}, we introduce three $\lambda$-strategies from the literature and propose five new ones, including the above-mentioned Global Std strategy. Given the theoretical foundations, in \textbf{Section \ref{sec:exp}}, we describe our experiment setup and then evaluate UCT with the different $\lambda$-strategies on six problems. Lastly, in \textbf{Section \ref{sec:future_work}}, we briefly summarize our findings and give an outlook for future work.

\section{Related work}
\label{sec:related_work}
Surprisingly, there is very little research conducted on dynamically choosing the exploration constant $\lambda$ in MCTS (or equivalently normalizing Q-values, see Section \ref{sec:method}) \cite{BrownePWLCRTPSC12} on-the-fly. Even less research has been conducted on scale-invariant exploration strategies. We suspect this to be the case since this is only a minor detail in MCTS, mostly not worthy of spending dedicated research effort on. Further, we also assume that there are numerous papers we are not aware of that employed some adaptive $\lambda$-strategy for their UCT. We assume this because authors may not have advertised their $\lambda$-strategy as it was only considered a minor implementation detail.

Tomáš Kozelek \cite{kozelek2009mcts} suggested two $\lambda$-strategies, one where $\lambda$ decays proportional to $\frac{1}{\sqrt{n}}$ and one where $\lambda$ is proportional to the empirical variance of the current node's returns. In both strategies, $\lambda$ is clipped within a fixed range. Both Auer et al. \cite{AuerCF02} and later Audibert et al. \cite{audibert} also proposed reward-variance-based $\lambda$-strategies called UCB-tuned and $\beta$-UCB. For each action, they use the variance of the state-action pair's returns instead of the entire node's returns. Yet another $\lambda$-strategy paradigm was introduced by Moerland et al. \cite{moehrland} who used a $\lambda$ value that directly reflects the size of the current node's subtree, which quantifies uncertainty, thus improving exploration.
Even without clipping, none of these strategies 
are independent of the scale of the Q-values.

One of the few documented scale-independent tuning approaches is choosing $\lambda = Q_a$ \cite{BallaF09,BonetG12,EyerichKH10} where $Q_a$ is the Q-value of the action $a$ under consideration at this node. However, a big flaw of this approach is that any actions with negative Q-values are always ignored no matter the visit count, when there is at least one action with a positive Q-value. This issue can be avoided by using $\lambda = |Q_a|$ as done by Anand et al in OGA-UCT \cite{OGAUCT}. However, this comes with the problem that there is a bias towards actions with $Q_a \ll 0$.
This difficulty can be circumvented with a $\lambda$-stratey proposed by Schrittwieser et al. \cite{SchrittwieserAH20} who choose $\lambda = Q^{\mathcal{T}}_{\text{max}} - Q^{\mathcal{T}}_{\text{min}}$ where $Q^{\mathcal{T}}_{\text{max}},Q^{\mathcal{T}}_{\text{min}}$ are the highest and lowest Q values of the current search tree $\mathcal{T}$. Yet another scale-independent approach that is also reward-variance-based is given by 
Gray et al.'s \cite{polyucb} strategy called \textit{Poly-UCB1 with derived C} that considers different confidence levels of the Student's t-distribution. In contrast to our Global Std method, Gray et al. build the variance from the returns of a single Q-value, while Global Std uses the variance of the different Q values. 

In a broader scope, 
there exists a plethora of techniques to improve MCTS \cite{BrownePWLCRTPSC12}, some of which directly target the tree policy, such as work by Galván et al. \cite{GalvanSA23}, who use evolutionary algorithms to find alternatives for the entire UCB formula or techniques that only partially affect the exploration term, such as work by
Sironi et al. \cite{SironiLLGBLW18} who propose and review several online parameter tuning techniques that interleave UCT iterations and parameter selection. Aside from MCTS, parameters in a more general reinforcement learning setting can also be scale-dependent, such as the temperature in Soft Actor Critic, for which automatic tuning methods exist \cite{sacext}. 

\section{Method}
\label{sec:method}
\subsection{Problem model and optimization objective}
Firstly, we use finite 
Stochastic Games (SG)
as the model for the environments/tasks/problems considered here, which UCT formally operates on. This subsequent definition is a slight modification of Shapley's original definition \cite{Shapley1953}. In the following, $\Delta (X) \subseteq \mathbb{R}^n$ denotes the probability simplex for any non-empty set $X$ with $|X| = n$.
\begin{definition}
    A \textit{SG} is an 8-tuple $M=(S,n_{\text{p}},\mu_0,T,I, \mathbb{A},\mathbb{P}_{\text{t}}, R)$ where the components are as follows:
    \begin{itemize}
        \item $S \neq \emptyset$ is the finite set of states, $n_{\text{p}} \in \mathbb{N}$ is the number of players, $\mu_0 \in \Delta(S)$ is the probability distribution for the initial state, and $T \subseteq S$ is the (possibly empty) set of terminal states.
         \item $I\colon (S \setminus T) \mapsto \{1,\dots,n_{\text{p}}\}$ maps each non-terminal state to which of the $n_{\text{p}}$ players is at turn.
        \item $\mathbb{A} \colon (S \setminus T) \mapsto A$ maps each state $s$ to the available actions $\emptyset \neq \mathbb{A}(s) \subseteq A$ at state $s$ where $|A| < \infty$. Again, we denote the set of permissible state-action pairs by $P_M = \{(s,a)\:|\: s \in (S \setminus T), a \in \mathbb{A}(s)\}$.
        \item $\mathbb{P}_{\text{t}}\colon P_M \mapsto \Delta(S )$ is the stochastic transition function where we use $\mathbb{P}_{\text{t}}(s^{\prime} |\: s,a)$ to denote the probability of transitioning from $s \in S$ to $s^{\prime} \in S$ after taking action $a \in \mathbb{A}(s)$ in $s$.
        \item $R \colon P_M \mapsto \mathbb{R}^{n_{\text{p}}}$ is the reward function that outputs the reward of each player.
    \end{itemize}
\end{definition}
In particular, note that
we assume  that the immediate reward deterministically depends on the current state and chosen action only, and that each state has exactly one player who may choose the next action. Furthermore, if we choose $n_{\text{p}} = 1$, then one obtains an MDP \cite{sutton2018reinforcement}.

From hereon, let $M = (S,\mu_0,\mathbb{A},\mathbb{P}, R, T)$ be a SG. 
The goal is to find an agent $\pi  \colon S \mapsto \Delta(A)$ that is modelled as a mapping from states to action distributions
such that $\pi$ maximizes the expected episode's return for player $1 \leq i \leq n_{\text{p}}$ where the (discounted) return for of episode $s_0,a_0,r_0, \dots, s_n,a_n,r_n,s_{n+1}$ with $s_{n+1} \in T$ is given by $\gamma^0 (r_0)_i + \ldots + \gamma^n (r_n)_i$.

\subsection{MCTS foundation and equivalences between normalization and $\lambda$-strategies}
The tree policy in UCT for fully expanded nodes is given by selecting the action with the highest UCB value, with random tie-breaking at each node. The UCB value at a node $\mathcal{N}$ of an action $a \in A$ with $N_a \in \mathbb{N}$ visits and an estimated Q-value $Q_a \in \mathbb{R}$ (i.e. accumulated reward divided by the number of visits) with $\mathcal{N}$ having $n \in \mathbb{N}$ total visits is given by
   $ Q_a + \lambda \cdot \sqrt{\frac{\log{n}}{N_a}}$,
where $\lambda \in \mathbb{R}^+$ is the exploration constant. In this paper, we only discuss strategies for picking $\lambda$, however, this is equivalent to any Q-value normalizing strategies, since it holds that
\begin{align}
    \underset{a \in A}{\textrm{argmax }} Q_a + \lambda \cdot \sqrt{\frac{\log{n}}{N_a}} &= \\
    \underset{a \in A}{\textrm{argmax }} \frac{Q_a - Q_{\text{lower}}}{Q_{\text{upper}} - Q_{\text{lower}}} + \sqrt{\frac{\log{n}}{N_a}} &
\end{align}

\noindent with $Q_{\text{lower}},Q_{\text{upper}} \in \mathbb{R}$ being some normalizing constants, $0 \neq \lambda = Q_{\text{upper}} - Q_{\text{lower}}$.

\subsection{Introduction of $\lambda$-strategies}
Instead of $\lambda$ being constant, we treat it as a function of the current search tree $\mathcal{T}$, the tree policy decision node $\mathcal{N}$, and the considered action $a$ to allow for information flow between all nodes and not just those on the same path to the root. 

Next, we denote the different variants for $\lambda \coloneqq \lambda(\mathcal{T},\mathcal{N},a)$ that we considered where we assume $\lambda$ to have the form $\lambda = C \cdot \hat{\lambda},\ C \in \mathbb{R}$. Even though this form still contains a parameter, this one will be by construction scale independent and less sensitive to the specific environment as we shall see later. After we introduce the $\hat{\lambda}$ variants, we will give additional reasoning for why this $C$ is necessary. The variants with the names $\{$Global Abs, \textbf{Global Std}, Layer Range, Layer Abs, Layer Std, Local Range, Local Std$ \}$ are introduced by us, while the others have already been used throughout the literature.
\\ \\
    \noindent \textbf{Vanilla UCT}: $\hat{\lambda} =\hat{\lambda}_0$ is constant. This strategy was originally proposed by Kocsis and Szepesvari\cite{KocsisS06} and does not perform any dynamic adjustment of $\hat{\lambda}$.

    \noindent \textbf{Global Range}: $\hat{\lambda}= Q^{\mathcal{T}}_{max} - Q^{\mathcal{T}}_{min}$ where $Q^{\mathcal{T}}_{max},Q^{\mathcal{T}}_{min}$ denote the maximum and minimum Q-value of all state-action pairs in $\mathcal{T}$. This strategy was proposed by Schrittwieser et al. \cite{SchrittwieserAH20} and does arguably not satisfy the no-computational-overhead requirement as one either has to regularly iterate the entire tree or rearrange a sorted list of all Q values to know the min and max Q value. However, we still included it for comparison purposes and as we shall see later it performs worse than our Global Std method.

    \hspace*{-0.4cm}\rule{0.495\textwidth}{0.44pt}

    \noindent \textbf{Global Abs}: $\hat{\lambda}$ is set to the average absolute Q-value of all state-action pairs in $\tau$.

    \noindent \textbf{Global Std}: $\hat{\lambda}= \sigma^{\mathcal{T}}$, where $\sigma^{\mathcal{T}}$ denotes the empirical standard deviation of the Q-values of all state-action pairs in $\mathcal{T}$. Hence, if $\mathcal{T}$ contains $n$ state-action pairs, $\sigma^{\tau}$ is the standard deviation of $n$ data points.

    \noindent \textbf{Layer Range}: $\hat{\lambda}= Q^{l}_{max} - Q^{l}_{min}$ where $Q^{l}_{max},Q^{l}_{min}$ denote the maximum and minimum Q-value of all state-action pairs in the same layer as $\mathcal{N}_a$. As with Global Range, this method also brings a computational overhead with it, however, as we shall see later, Global Std outperforms it nonetheless. 

    \noindent \textbf{Layer Abs}: $\hat{\lambda}$ is set to the average absolute value of all state-action pairs in the same layer as $\mathcal{N}$.

    \noindent \textbf{Layer Std}: $\hat{\lambda}= \sigma^{l}$, where $\sigma^{l}$ denotes the empirical standard deviation of the Q-values of all state-action pairs in the same layer as $\mathcal{N}_a$.

    \noindent \textbf{Local Range}: $\hat{\lambda}= Q^{\mathcal{N}}_{max} - Q^{\mathcal{N}}_{min}$ where $Q^{\mathcal{N}}_{max},Q^{\mathcal{N}}_{min}$ denote the maximum and minimum Q-value of all state-action pairs of $\mathcal{N}$.

    \noindent \textbf{Local Std}: $\hat{\lambda}= \sigma^{\mathcal{N}}$, where $\sigma^{\mathcal{N}}$ denotes the empirical standard deviation of the Q-values of all state-action pairs of $\mathcal{N}$.

    \hspace*{-0.4cm}\rule{0.495\textwidth}{0.44pt}

     \noindent \textbf{Local Abs Q}: $\hat{\lambda}= |Q_a|$ where $Q_a$ denotes the Q-value of action $a$ at $\mathcal{N}$. This strategy was used by Anand et al. \cite{OGAUCT}.

    \noindent \textbf{Local Q}: $\hat{\lambda} = Q_a$ where $Q_a$ denotes the Q-value of action $a$ at $\mathcal{N}$. This strategy is the most common in literature \cite{BallaF09,BonetG12,EyerichKH10}.
    
    \noindent \textbf{Poly-UCB1}: $\hat{\lambda} = \frac{\sigma_a \cdot \varphi(N_a)}{\log n}$ where $n$ is the number of visits of node $\mathcal{N}$, $N_a$ is the number of times action $a$ has been visited at node $\mathcal{N}$, $\varphi(N_a)$ is the $99\%$ quantile of the Student's t-distribution with $N_a - 1$ degrees of freedom, and 
    $\sigma_a$ is the empirical standard deviation for the state-action pair $(\mathcal{N},a)$. Note that $\sigma_a$ is not the same as $\sigma^{\mathcal{N}}$ as the latter considers different Q values, while $\sigma_a$ considers individual returns of a single Q-value.

\subsection{Theoretical discussion of the $\lambda$-strategies}

\noindent\textbf{Why the parameter} $\bm{C}$\textbf{ is necessary}:
Even though the above-mentioned $\lambda$-strategies are reward scale independent, some problems may favor more explorative tree-policies while other favor more exploitative ones. Hence, we still need the parameter $C$ to control this. 

Also note that since the different scopes (e.g. global, layer, local) form a hierarchy of nodes, it follows that $Q_{max}^{\mathcal{T}} - Q_{min}^{\mathcal{T}} \geq Q_{max}^l - Q_{min}^l \geq Q_{max}^{\mathcal{N}} - Q_{min}^{\mathcal{N}}$. Even though all of these compute the same quantity in principle, their scale differs which can be fixed by the multiplication with $C$ depending on the strategy used. 

\noindent\textbf{Fulfilment of scale independence}:
Observe that these $\lambda$-strategies satisfy the scale independence condition posed in the introduction as
all presented variants for $\lambda$ use functions $f$ (namely: $| \cdot |, \sigma, \max - \min, \textrm{id}$) with the homogeneity property that for any $\mu > 0$ and input $x$ it holds that $f(\mu x) = \mu f(x)$.

The validity of achieving reward scale independence by choosing homogenous function is supported by reconsidering the convergence analysis of the original UCT paper \cite{KocsisS06} where Kocsis
and Szepesv{\'{a}}ri assumed all cumulative rewards to lie in the interval $[0,1]$. If one were to preserve their results for arbitrary bounds $[a,b]$, one simply needs to scale the exploration constant with $b-a$. This is because their analysis builds on bounds obtained by Hoeffding's inequality \cite{hoeffding1963probability} which states for any independent random variables $a \leq X_1,\dots,X_n \leq b$ and any $t>0$ it holds that
\begin{equation}
    \mathbb{P}\left(\sum_{i=1}^n (X_i - \mathbb{E}[X_i]) \geq t\right) \leq e^{-\frac{2nt^2}{(b-a)^2}}.
\end{equation}
Without further specifying what the $X_i$ are, Kocsis
and Szepesv{\'{a}}ri insert the exploration constant for $t$. Hence, to obtain the same bounds as for $(a,b) = (0,1)$ one has to replace $t$ (and subsequently the exploration constant) with $(b-a)t$.

\noindent\textbf{Fulfilment of low computational overhead}:
The $\lambda$-strategies introduced here, except for the range strategies, also fulfill the remaining two conditions posed in the introduction. For calculating the standard deviation, one only needs to
keep track of the number of Q-values in consideration, their
sum, and their squared sum and then make use of the identity
$\textrm{Var}(X) = \mathbb{E}[X^2] - \mathbb{E}[X]^2$ for a random variable $X$. The bookkeeping needed for all strategies can be computed
during the backpropagation phase of MCTS: The old Q- and squared Q-value is subtracted from the statistics under consideration and the updated ones are added back.

\noindent\textbf{Additive invariance of std and range}:
Lastly, note that the std and range strategies do not change their output when a constant is added to all Q-values their computation is based on. This property is desirable for MDPs with fixed episode lengths as the optimal policy in these cases is invariant to adding a constant to the reward function.

\noindent\textbf{Convergence of std strategies}: While both the std and range strategies fulfill all of the above-mentioned properties, range strategies are not guaranteed to converge. Imagine an MCTS search tree where the Q-values are sampled from a Gaussian. While the std strategies will eventually converge to a unique value, all range strategies will diverge to positive infinity with an increase in the search tree size as more outliers are sampled. In general, the range strategies are vulnerable to outlier values.

\begin{figure}[ht]
  \centering

  \begin{subfigure}[b]{0.22\textwidth}
    \centering
    \includegraphics[width=\textwidth]{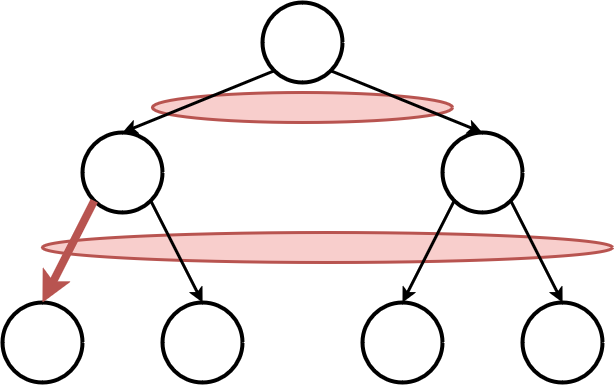}
    \caption{Used by: Global Range, Global Std, Global Abs}
  \end{subfigure}
  \hfill
  \begin{subfigure}[b]{0.22\textwidth}
    \centering
    \includegraphics[width=\textwidth]{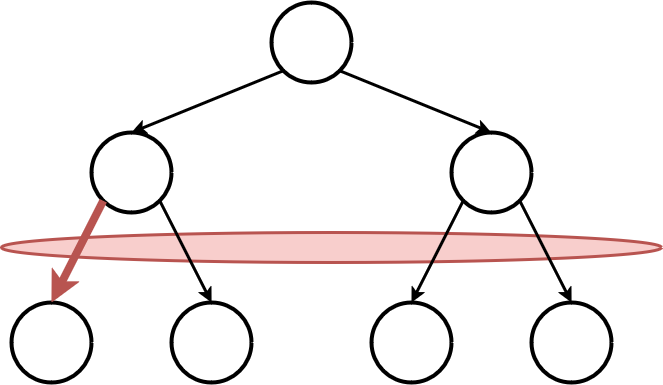}
    \caption{Used by: Layer Range, Layer Std, Layer Abs}
  \end{subfigure}
  \hfill
  \begin{subfigure}[b]{0.22\textwidth}
    \centering
    \includegraphics[width=\textwidth]{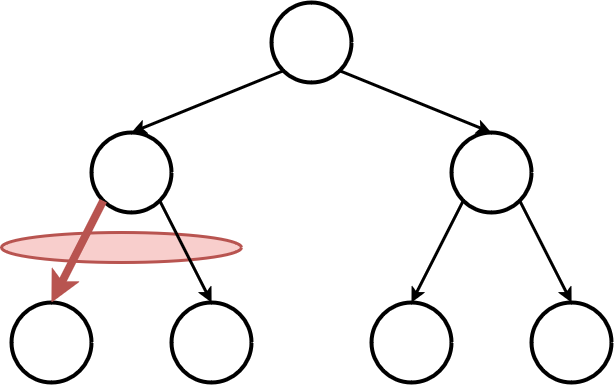}
    \caption{Used by: Local Range, and Local Std}
  \end{subfigure}
  \hfill
    \begin{subfigure}[b]{0.22\textwidth}
    \centering
    \includegraphics[width=\textwidth]{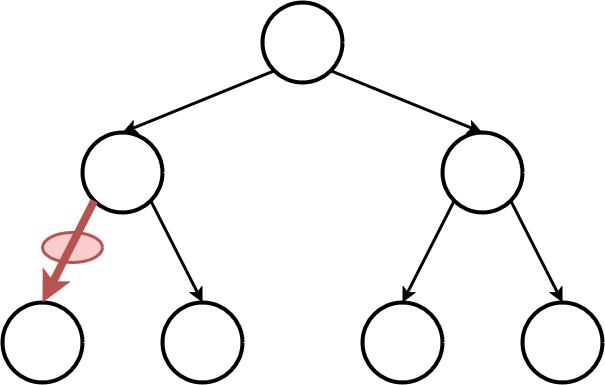}
    \caption{Used by: Local Abs, Local Q, Poly-UCB1}
  \end{subfigure}

  \caption{A visualization of which state-action pairs are taken into consideration when determining the exploration factor $\lambda$ for the bottom-left state-action pair that is marked in red. The four images all depict a search tree with five nodes where arrows represent deterministic actions. The state-action pairs used to determine the corresponding exploration constant all intersect a red ellipse.
  Each subfigure's caption lists the $\lambda$-strategies that use the marked set of state-action pairs for the $\lambda$ calculation.}
  \label{fig:localities}
\end{figure}

\noindent\textbf{Localities}:
Most methods can be grouped by their locality, i.e. global, versus layerwise, versus local. Each comes with a different theoretical tradeoff: The larger the scope, the more values can be aggregated to compute the desired quantity (i.e., standard deviation, absolute value, or range) but the less accurate these estimates are for the current node $\mathcal{N}$ in question. Fig.~\ref{fig:localities} visualizes the different localities.

\section{Experiments}
\label{sec:exp}
\begin{figure*}[ht]
  \centering
    \includegraphics[width=1\textwidth]{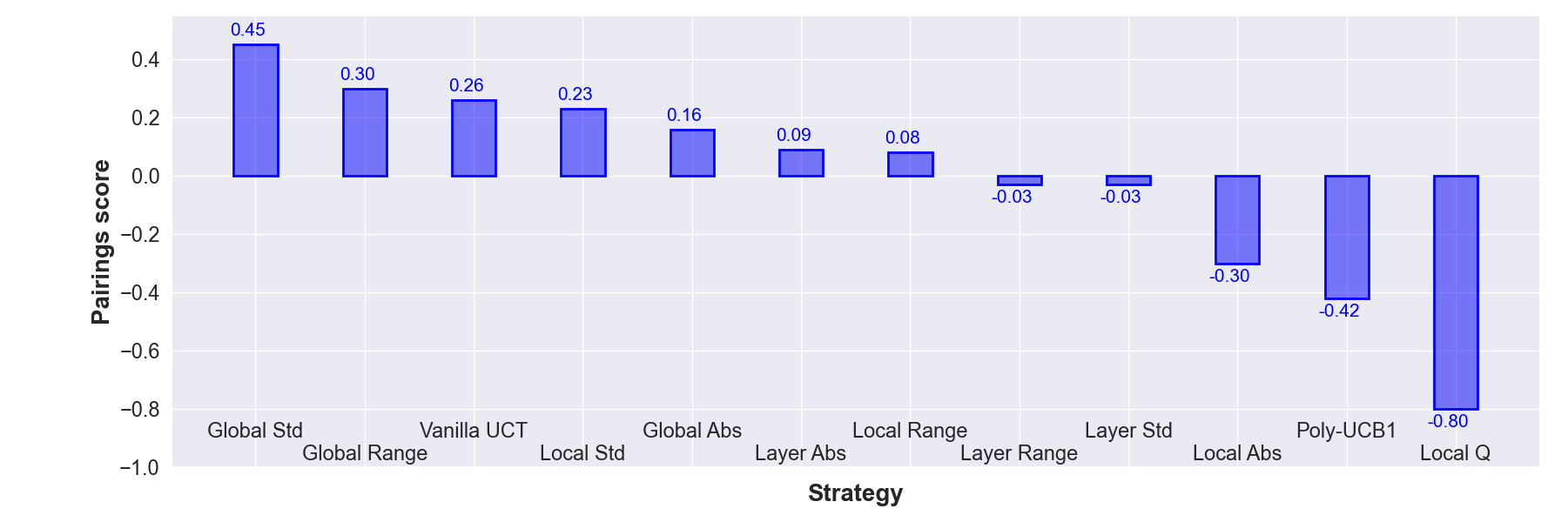}
  \caption{The normalized pairings score for all considered $\lambda$-strategies. The score was constructed by considering all iteration budgets and all environments. Per pairing, the maximum performance over all $C$ values (i.e. $\{0.125,0.25,0.5,1,2,4,8,16,32,64\}$ for all strategies and additionally $256$ and $1000$ for Vanilla UCT) was used per iteration-environment pair. One of our methods, Global Std, performs best overall.}
  \label{fig:score_over_all}
\end{figure*}

\begin{figure}[h]
  \centering
    \includegraphics[width=0.5\textwidth]{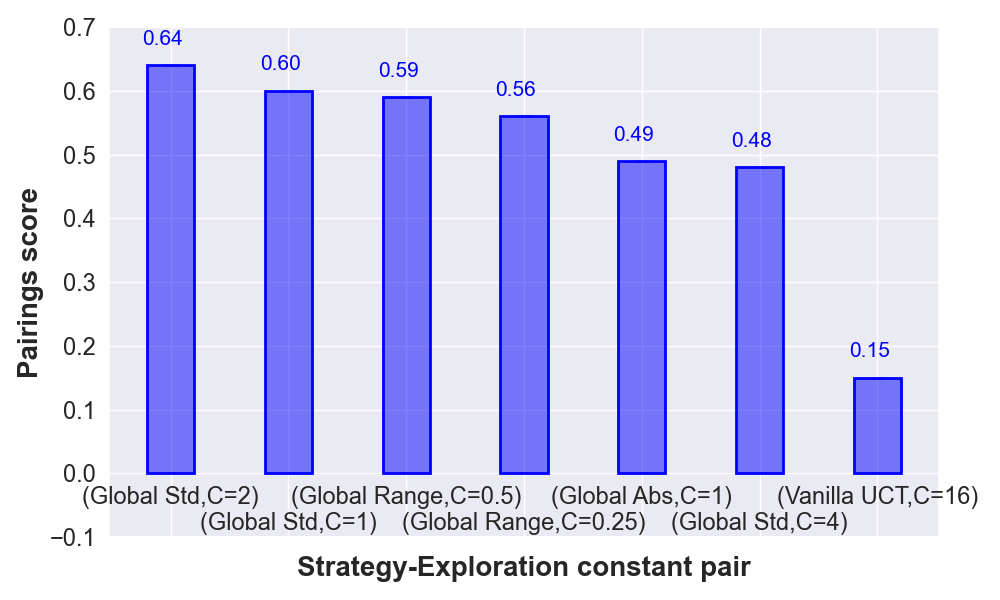}
  \caption{The best 6 normalized pairings score for all considered $\lambda$-strategy and exploration constant $C$ pairs as well the best UCT parameter. The score was constructed by considering all iteration budgets and all environments. One of our methods, Global Std with $C=2$, performs best overall, while standard UCT is far worse than any of these methods, proving that scale-invariant methods are necessary. The score for all $\lambda$-strategy-$C$-pairs is found in the last column of Tab.~\ref{tab:scores_strategy_and_c} in the supplementary materials.}
  \label{fig:score_gen}
\end{figure}

\subsection{Experiment setup}
\label{sec:exp_setup}
In this section, we describe the experiment setup for the subsequent experiments.

\noindent \textbf{Problem models:}
For this paper, we ran our experiments on 17 MDPs and 11 two-player games, all of which are either from the International Probabilistic Planning Conference \cite{grzes2014ippc}, or are well-known board or strategy games. 

\noindent\textit{List of MDPS}: Academic Advising, Cooperative Recon, Crossing Traffic, Earth Observation, Elevators, Game of Life, Manufacturer, Navigation, Push Your Luck, Racetrack, Red Finned Blue Eye, Sailing Wind, Skill Teaching, Saving, SysAdmin, Tamarisk, Traffic, Triangle Tireworld, Wildfire, and Wildlife Preserve.

\noindent \textit{List of two-player games}: Chess, Constrictor, Connect 4, Numbers Race, Othello, Quarto, Pylos, Capture the Flag, Kill the King, Pusher, and Tic Tac Toe.

In the supplementary materials in Section \ref{sec:problem_list}, we provide a description of all the above-mentioned problems.
All experiments were run on the finite horizon versions of the considered MDPs with a default horizon of 50 steps and 200 for the two-player SGs with a planning horizon of 50 and a discount factor $\gamma=1$. On SGs, agents were evaluated versus the fixed 500 iterations Global Std agent with $C=4$.
In particular, these tasks cover different reward scales, for example the two-player zero-sum games' episode returns are either 1,-1, or 0, Game of Life's episodes' returns lie in the positive 3-digit region while Sailing Wind's returns are each within the negative 2-digit region.

\noindent \textbf{Evaluation:}
Each data point that we denote in the remaining sections of this paper (e.g. agent returns) is the average of at least 2000 runs. For the two-player games, we ran 2000 games where the agent-to-be-evaluated has the first move and 2000 where the agent does not.
Whenever we denote a confidence interval for a data point, then this is always a confidence interval with a confidence level of 99\% provided by $\approx 2.33$ times the standard error.

\noindent \textbf{Normalized pairings score}: Mostly, performance margins will be rather small. To be able to appropriately rank the $\lambda$-strategies without manually reviewing over 1000 individual performances, we will later construct so-called 
\textit{normalized pairings-score matrices} and the \textit{normalized pairings score}  that is constructed as follows. Let $\{\pi_1,\dots,\pi_n\}$ be $n$ agents (e.g., parameter-optimized $\lambda$-strategies, or individual $\lambda$-strategy-parameter pairs) where each agent was evaluated on $m$ tasks (later, a task will be a given MCTS iteration budget and an environment). The corresponding normalized pairings score matrix is of size $n \times n$ and the entry $(i,j)$ is equal to the number of tasks where $\pi_i$ performed better than agent $\pi_j$  subtracted by the number of times it performed worse, divided by $m$.
The normalized pairings score $s_i,1 \leq i \leq n$ is given by averaging over the $i$-th row when excluding the $i$-th column. 

\noindent \textbf{Reproducibility:}
For reproducibility, we released our implementation which is available at \url{https://github.com/codebro634/DynamicExplorationFactorUCT}.  Our code was compiled with g++ version 13.1.0 using the -O3 flag (i.e. aggressive optimization). For reproducibility purposes, we seeded all components of our program that require RNG. The RNG engine was reset for every Agent-Domain-Map triplet with the seed $42$.

\subsection{Experimental results}
Using the previously described setup, we measured the mean episode return for each $\lambda$-strategy-environment pair using $n \in \{100,500,2500\}$ MCTS iterations and $C \in \{0.125,0.25,0.5,1,2,4,8,16,32,64\}$ for all $\lambda$-strategies except for Vanilla UCT for which we used $C \in \{0.125,0.25,0.5,1,2,4,8,16,32,64, 256, 1000\}$ to account for environments which high-magnitude reward scales. In the following, we will only be presenting a summary of these experiments. The ful data table(s) can be found in the supplementary materials in Subsection \ref{sec:ful_data_tables}.

\noindent\textbf{Investigating peak performances:} First, we tested which of the strategies introduced here performs best when considering the best performing $C$ value for each iteration budget-model pair. Bar chart~\ref{fig:score_over_all} shows the normalized pairings score (see Section \ref{sec:exp_setup}) for all 12 parameter-optimized $\lambda$-strategies across all iteration budgets. 
Firstly, our newly proposed strategy, Global Std, performed best overall, followed by a significant gap by Global range. This lead is kept for all the iteration budgets considered here. In the supplementary materials, we list the normalized pairing scores for each iteration budget in Tab.~\ref{tab:optimized_by_iteration}. Furthermore, in Tab.~\ref{tab:score_matrix_all}, we denote the normalized pairings score matrix for all iteration budgets combined. Tables~\ref{tab:score_matrix_100},~\ref{tab:score_matrix_500}, and ~\ref{tab:score_matrix_2500} show the normalized pairings score matrices for the individual iteration budgets.

One can also make another key observation from this bar chart: Global strategies, that use all Q nodes as samples, perform best overall, while the most local strategies, namely Local Q, Poly-UCB1, and Local Abs, perform worst overall.

Lastly, it has to be noted that while Global Std's pairings score far exceeds that of Vanilla UCT, this does not take the magnitude of performance improvements into account. Tab.~\ref{tab:performances_versus_optim} in the supplementary materials shows the exact performances for each individual environment of Global Std, Vanilla UCT, and the strategy that performed best in the corresponding environment. Expectedly, the relative performance improvement over Vanilla UCT is rather small (on average 4\%), as optimizing Vanilla UCT per environment does not have any issues with varying reward scales. 

\noindent\textbf{Investigating generalization capabilities:} Next, we investigate the generalization capabilities to determine which single parameter combination performs best overall. Bar chart~\ref{fig:score_gen} shows the top 6 normalized pairings scores for all $\lambda$-strategy, exploration constant $C$ pairs as well as the best performing Vanilla UCT variant for comparison. Though with a smaller lead, our method Global Std performs best overall using the parameter $C=2$. The second-best $\lambda$-strategy is Global Range with $C=0.5$. Note that the best Vanilla UCT variant ($C=16$) performs far worse than these scale-invariant methods, showing that fixed-parameter UCT can not deal with varying reward scales, as Vanilla UCT dropped from third place in the parameter-optimized setting to some middle-of-the-pack place in the generalization setting. To emphasize the reward scale sensitivity, Fig.~\ref{fig:varying_reward_scales} in the supplementary materials shows the performances of Global Std and Vanilla UCT with fixed $C$ values but varying reward scales for the environments considered here.

Again, the results are not dependent on the concrete MCTS iteration budget: The parameter $C=2$ for Global Std performs best for all iteration budgets. Furthermore, except for the 100 setting where Global Std is nearly equal to Global range, Global Std always performs best. In the supplementary materials in Tab.~\ref{tab:scores_strategy_and_c}, we list the normalized pairings score for all iteration budgets.

Additionally, in contrast to the parameter-optimized setting, a single $C$ value for Vanilla UCT is rarely adapted to the problem's reward scale. Tab.~\ref{tab:performances_versus_gen} in the supplementary materials shows the exact performances on each environment of Global Std using $C=2$ and Vanilla UCT using $C=16$. In this case, Global Std has an on average 40\% higher performance than Vanilla UCT.

Lastly, to test whether the performance improvements Global Std has over Vanilla UCT stem solely from Global Std being reward scale invariant,
the pairings score for both the parameter-optimized and single-parameter setting was calculated when restricted to the two-player zero-sum games, which all have the same reward scale. The results are shown in Fig.~\ref{fig:score_over_all_2p} and Fig.~\ref{fig:score_gen_2p} in the supplementary materials. 
Unsurprisingly, Vanilla UCT performs far better in this setting; however, in direct comparison, Global Std still has a notable lead over Vanilla UCT. This suggests that Global Std has effects beneficial to the performance that go beyond adapting the reward scale. We believe the global Q variance is a proxy for uncertainty and thus the amount of exploration required.

\section{Conclusion and Future Work}
\label{sec:future_work}
In this paper, we evaluated 12 different $\lambda$-strategies (5 from the literature and 7 from us) on a plethora of sequential-decision making problems with varying reward scales, showing that Vanilla UCT (i.e., using a fixed exploration constant) is not only highly scale dependent, even with a properly adjusted $\lambda$, Vanilla UCT is still outperformed by some of the suggested strategies. In particular, one of our newly proposed strategies, namely Global Std using $C=2$, generalized best across the environments considered here and also has the best overall parameter-optimized performance, decisively beating Vanilla UCT in both regards.
Finally, we came to the conclusion that the Global Std strategy is a good, easy-to-implement replacement for Vanilla UCT.

For future work, we think that one can expand this work in two dimensions. Firstly, propose and evaluate more $\lambda$-strategies, for example, by taking a convex combination of the already introduced strategies or dynamically switching between strategies, as though Global Std performs best overall, it doesn't do so for every single environment. Secondly, the evaluation may be conducted on a much larger suite of environments. For example, we did not include partially observable or deterministic single-agent environments, and we are far from exhausting the IPPC problem list.

\bibliographystyle{plain}
\bibliography{references}

\clearpage
\onecolumn

\section*{Supplementary materials}
\label{sec:appendix}

\subsection{Performances for fixed environments with varying reward scales}

\begin{figure}[H]
\centering

\begin{minipage}{0.32\textwidth}
\centering
\includegraphics[width=\linewidth]{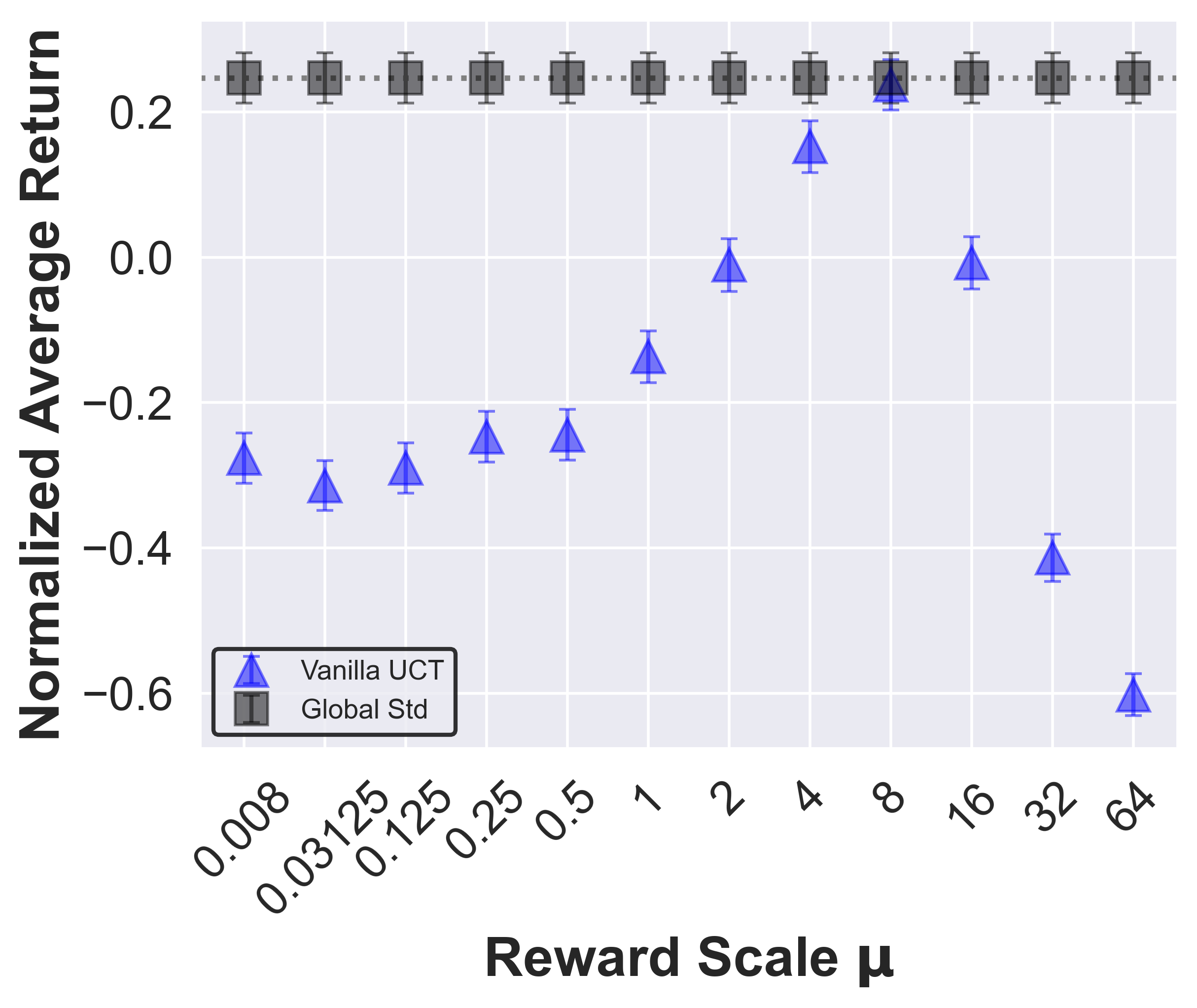}
\caption*{(a) Connect 4}
\end{minipage}
\hfill
\begin{minipage}{0.32\textwidth}
\centering
\includegraphics[width=\linewidth]{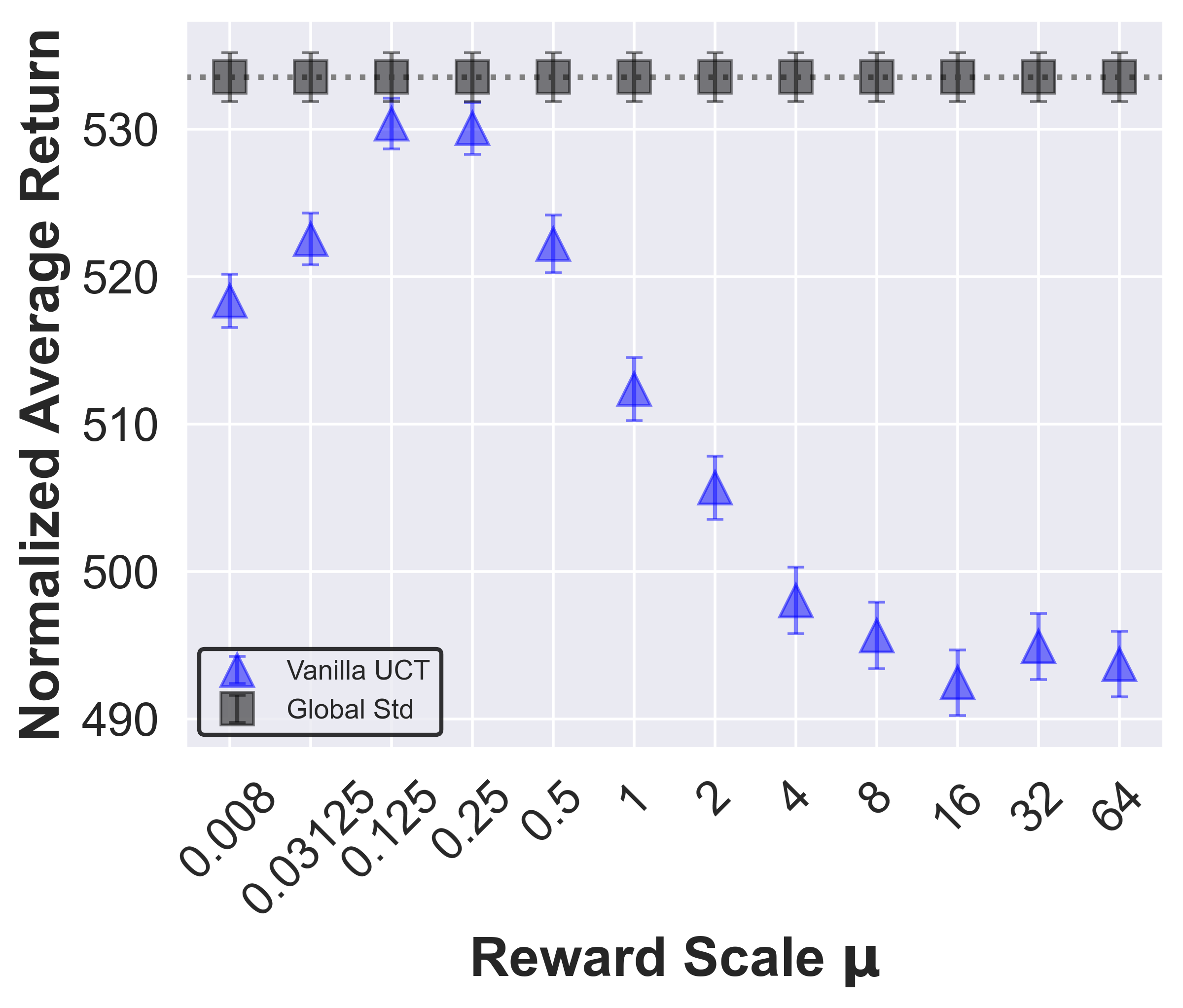}
\caption*{(b) Game of Life}
\end{minipage}
\hfill
\begin{minipage}{0.32\textwidth}
\centering
\includegraphics[width=\linewidth]{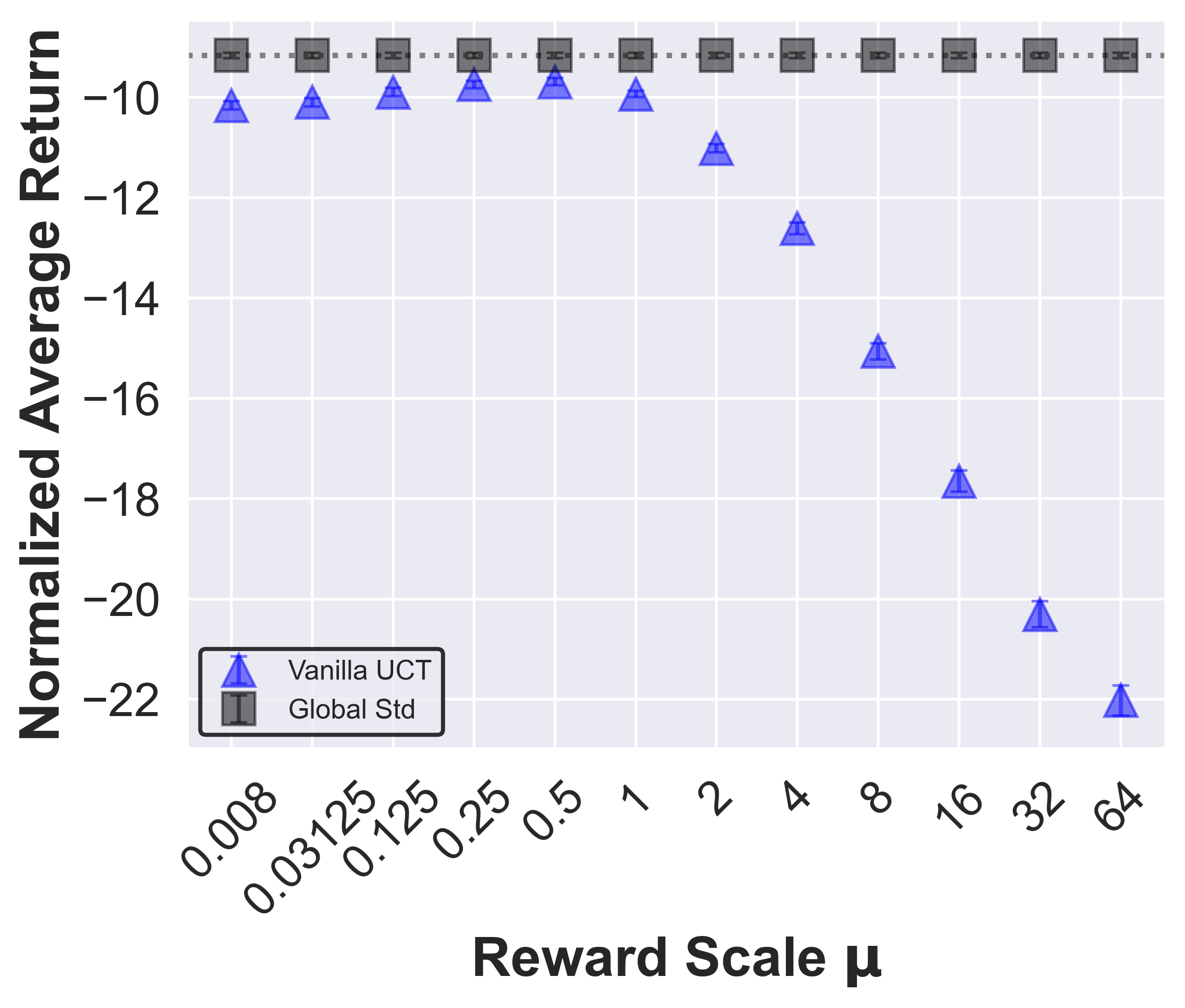}
\caption*{(c) Racetrack}
\end{minipage}

\caption{ Global Std is invariant to the reward scale of the environment, and scaling an environment's reward function by $\mu$ is equivalent to dividing the exploration constant by $\mu$ in Vanilla UCT. Using this fact, the hypothetical performances of Global Std and Vanilla UCT for 500 iterations are shown for three different environments when their respective reward function is scaled by a constant $\mu$. To better compare performances, if the reward function is scaled by $\mu$, then the average return is divided by $\mu$. For all environments, the value $C=8$ was used for Vanilla UCT, and for Global Std, the best performing $C$ value for the non-scaled environment (i.e., $\mu=1$) was used. Note that by construction, Global Std is invariant to the reward scale, while any fixed exploration constant, as in Vanilla UCT, is highly scale dependent.
}
\label{fig:varying_reward_scales}

\end{figure}

\subsection{Pairings scores for the two-player games}

\begin{figure}[H]
  \centering
    \includegraphics[width=0.9\textwidth]{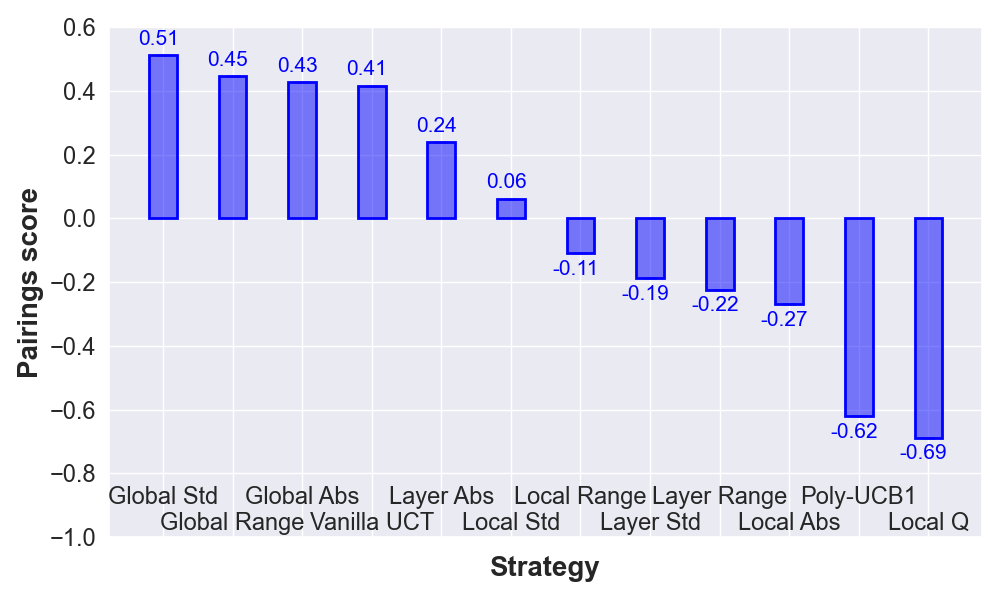}
  \caption{The normalized pairings score for all considered $\lambda$-strategies. The score was constructed by considering all iteration budgets and all \textbf{two-player games}. Per pairing, the maximum performance over all $C$ values (i.e. $\{0.125,0.25,0.5,1,2,4,8,16,32,64\}$ for all strategies and additionally $256$ and $1000$ for Vanilla UCT) was used per iteration-environment pair. One of our methods, Global Std, performs best overall.}
  \label{fig:score_over_all_2p}
\end{figure}

\begin{figure}[H]
  \centering
    \includegraphics[width=0.7\textwidth]{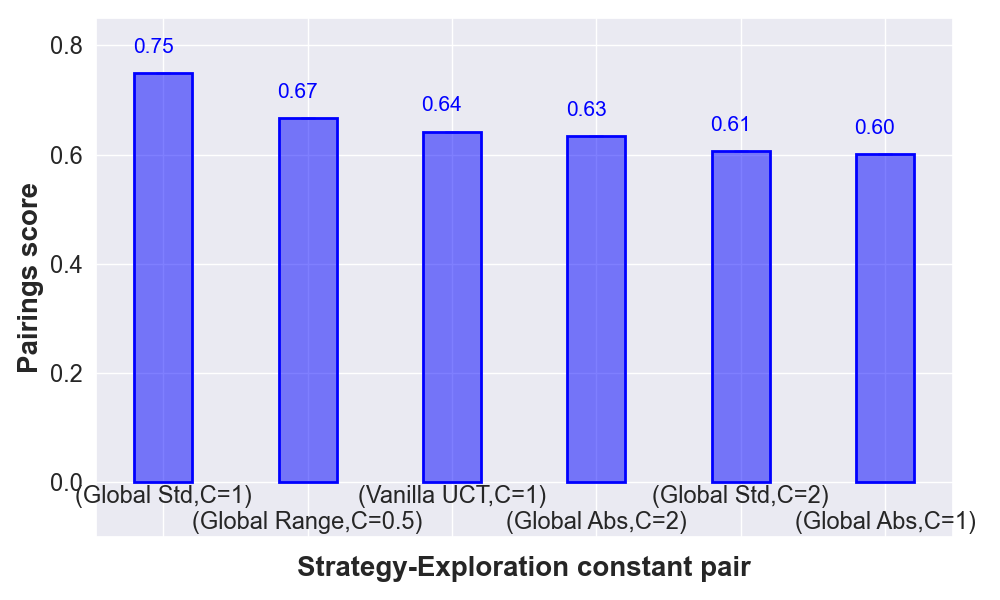}
  \caption{The best 6 normalized pairings score for all considered $\lambda$-strategy and exploration constant $C$ pairs. The score was constructed by considering all iteration budgets and all \textbf{two-player games}. One of our methods, Global Std with $C=1$, performs best overall, showing that Global Std even yields a small performance boost over Vanilla UCT in a constant reward scale setting.}
  \label{fig:score_gen_2p}
\end{figure}

\subsection{Parameter-optimized performances for each environment}

\begin{table}[h!]
\centering
\scalebox{0.75}{ 
\begin{tabular}{lcccccccccccc}
\toprule
 & \multicolumn{4}{c}{100 iterations} & \multicolumn{4}{c}{500 iterations} & \multicolumn{4}{c}{2500 iterations} \\
\cmidrule(lr){2-5} \cmidrule(lr){6-9} \cmidrule(lr){10-13}
Domain & Vanilla UCT & Global Std& Best & Best strategy & Vanilla UCT & Global Std& Best & Best strategy & Vanilla UCT & Global Std& Best & Best strategy \\
\midrule
Academic Advising & $-106.87 $ & $-105.29 $ & $-105.29 $ & \textbf{Global Std} & $-75.54 $ & $-72.76 $ & $-71.45 $ & Local Std& $-64.09 $ & $-62.67 $ & $-62.17 $ & Local Range\\
Connect4 & $-0.55 $ & $-0.58 $ & $-0.55 $ & Vanilla UCT & $0.24 $ & $0.25 $ & $0.25 $ & \textbf{Global Std} & $0.74 $ & $0.73 $ & $0.74 $ & Global Range\\
Chess & $-0.44 $ & $-0.39 $ & $-0.34 $ & Local Q & $0.64 $ & $0.65 $ & $0.67 $ & Global Abs & $0.99 $ & $0.99 $ & $1.00 $ & Layer Abs\\
Constrictor & $-0.38 $ & $-0.39 $ & $-0.37 $ & Global Abs & $0.05 $ & $0.04 $ & $0.06 $ & Global Abs & $0.38 $ & $0.38 $ & $0.41 $ & Global Range\\
Crossing Traffic & $-24.84 $ & $-24.72 $ & $-24.46 $ & Poly-UCB1 & $-25.38 $ & $-25.41 $ & $-24.42 $ & Local Std& $-24.63 $ & $-24.90 $ & $-24.21 $ & Poly-UCB1\\
CaptureTheFlag & $0.00 $ & $0.00 $ & $0.00 $ & Layer Abs & $0.00 $ & $0.00 $ & $0.00 $ & Layer Abs & $0.00 $ & $0.00 $ & $0.00 $ & \textbf{Global Std}\\
Earth Observation & $-14.21 $ & $-13.97 $ & $-13.49 $ & Local Std& $-8.53 $ & $-8.65 $ & $-8.50 $ & Poly-UCB1 & $-7.91 $ & $-7.79 $ & $-7.61 $ & Poly-UCB1\\
Game of Life & $484.55 $ & $486.76 $ & $488.48 $ & Poly-UCB1 & $530.36 $ & $533.52 $ & $533.52 $ & \textbf{Global Std} & $568.00 $ & $569.28 $ & $569.93 $ & Local Std\\
KillTheKing & $-0.16 $ & $-0.14 $ & $-0.14 $ & \textbf{Global Std} & $0.01 $ & $0.02 $ & $0.02 $ & \textbf{Global Std} & $0.14 $ & $0.15 $ & $0.15 $ & \textbf{Global Std}\\
Manufacturer & $-1586.81 $ & $-1564.31 $ & $-1553.79 $ & Poly-UCB1 & $-1329.48 $ & $-1325.67 $ & $-1325.67 $ & \textbf{Global Std} & $-1106.05 $ & $-1107.08 $ & $-1106.05 $ & Vanilla UCT\\
Navigation & $-29.78 $ & $-29.91 $ & $-29.42 $ & Layer Abs & $-23.38 $ & $-23.55 $ & $-23.23 $ & Layer Abs & $-21.57 $ & $-21.18 $ & $-19.95 $ & Poly-UCB1\\
NumbersRace & $-0.06 $ & $-0.06 $ & $-0.00 $ & Local Q & $0.83 $ & $0.83 $ & $0.83 $ & Vanilla UCT & $0.97 $ & $0.97 $ & $0.97 $ & Global Abs\\
Othello & $-0.46 $ & $-0.46 $ & $-0.44 $ & Global Range & $0.38 $ & $0.37 $ & $0.38 $ & Vanilla UCT & $0.88 $ & $0.88 $ & $0.88 $ & Vanilla UCT\\
Pusher & $-0.00 $ & $0.00 $ & $0.00 $ & Layer Abs & $0.00 $ & $0.00 $ & $0.00 $ & Vanilla UCT & $0.00 $ & $0.00 $ & $0.00 $ & Global Range\\
Push Your Luck & $54.19 $ & $54.05 $ & $54.31 $ & Local Std& $55.55 $ & $55.49 $ & $55.55 $ & Vanilla UCT & $54.96 $ & $54.90 $ & $55.98 $ & Poly-UCB1\\
Pylos & $-0.46 $ & $-0.47 $ & $-0.46 $ & Vanilla UCT & $0.27 $ & $0.26 $ & $0.28 $ & Global Range & $0.69 $ & $0.69 $ & $0.70 $ & Global Range\\
Quarto & $-0.30 $ & $-0.31 $ & $-0.30 $ & Vanilla UCT & $0.29 $ & $0.23 $ & $0.29 $ & Vanilla UCT & $0.51 $ & $0.49 $ & $0.52 $ & Global Range\\
Cooperative Recon & $5.15 $ & $5.18 $ & $5.37 $ & Global Abs & $6.37 $ & $6.40 $ & $6.40 $ & \textbf{Global Std} & $12.79 $ & $11.63 $ & $12.79 $ & Vanilla UCT\\
Racetrack & $-13.88 $ & $-13.79 $ & $-13.79 $ & \textbf{Global Std} & $-9.69 $ & $-9.16 $ & $-9.16 $ & \textbf{Global Std} & $-8.33 $ & $-8.18 $ & $-8.12 $ & Local Range\\
SysAdmin & $327.42 $ & $327.08 $ & $328.46 $ & Layer Range & $381.99 $ & $382.79 $ & $383.96 $ & Layer Std& $402.23 $ & $402.48 $ & $403.66 $ & Local Range\\
Saving & $44.42 $ & $44.34 $ & $44.56 $ & Global Range & $49.89 $ & $50.18 $ & $50.18 $ & \textbf{Global Std} & $52.52 $ & $52.25 $ & $52.52 $ & Vanilla UCT\\
Skills Teaching & $-109.05 $ & $-107.20 $ & $-100.68 $ & Global Range & $27.03 $ & $26.34 $ & $28.27 $ & Global Range & $60.21 $ & $70.82 $ & $73.91 $ & Local Range\\
Sailing Wind & $-81.09 $ & $-80.49 $ & $-80.42 $ & Local Std& $-66.00 $ & $-65.26 $ & $-65.05 $ & Poly-UCB1 & $-62.02 $ & $-61.87 $ & $-61.28 $ & Local Abs\\
Tamarisk & $-843.70 $ & $-842.21 $ & $-840.26 $ & Local Std& $-612.42 $ & $-612.12 $ & $-607.43 $ & Layer Range & $-522.57 $ & $-520.38 $ & $-516.79 $ & Local Range\\
Traffic & $-22.08 $ & $-21.83 $ & $-21.68 $ & Poly-UCB1 & $-15.38 $ & $-15.02 $ & $-15.02 $ & \textbf{Global Std} & $-12.72 $ & $-12.76 $ & $-12.67 $ & Local Std\\
Triangle Tireworld & $50.23 $ & $49.45 $ & $50.23 $ & Vanilla UCT & $76.58 $ & $77.29 $ & $77.29 $ & \textbf{Global Std} & $82.36 $ & $82.59 $ & $82.59 $ & \textbf{Global Std}\\
TicTacToe & $-0.12 $ & $-0.13 $ & $-0.12 $ & Global Range & $0.01 $ & $0.01 $ & $0.02 $ & Global Range & $0.05 $ & $0.05 $ & $0.05 $ & Vanilla UCT\\
Wildlife Preserve & $1201.99 $ & $1197.90 $ & $1209.23 $ & Poly-UCB1 & $1352.84 $ & $1354.70 $ & $1360.14 $ & Poly-UCB1 & $1377.81 $ & $1378.01 $ & $1378.01 $ & \textbf{Global Std}\\

\bottomrule
\end{tabular}
}
\caption{Performance comparisons between the parameter-optimized versions of Vanilla UCT, Global Std, and the best $\lambda$-strategy out of all $\lambda$-strategies considered here. Global Std has an average 4\% performance increase over Vanilla UCT in this setting.}
\label{tab:performances_versus_optim}
\end{table}

\subsection{Single parameter performances for each environment}

\begin{table}[H]
\centering
\scalebox{0.9}{ 
\begin{tabular}{lcccccc}
\toprule
 & \multicolumn{2}{c}{100 iterations} & \multicolumn{2}{c}{500 iterations} & \multicolumn{2}{c}{2500 iterations} \\
\cmidrule(lr){2-3} \cmidrule(lr){4-5} \cmidrule(lr){6-7}
Domain & Vanilla UCT ($C=16$) & Global Std($C=2$) & Vanilla UCT ($C=16$) & Global Std($C=2$) & Vanilla UCT ($C=16$) & Global Std($C=2$) \\
\midrule
Academic Advising & $-108.04 $ & $-107.91 $ & $-84.23 $ & $-74.09 $ & $-77.90 $ & $-63.03 $\\
Connect4 & $-0.82 $ & $-0.61 $ & $-0.24 $ & $0.14 $ & $0.22 $ & $0.66 $\\
Chess & $-0.56 $ & $-0.48 $ & $-0.10 $ & $0.15 $ & $0.71 $ & $0.98 $\\
Constrictor & $-0.43 $ & $-0.39 $ & $-0.09 $ & $0.04 $ & $0.08 $ & $0.37 $\\
Crossing Traffic & $-25.99 $ & $-25.71 $ & $-26.02 $ & $-26.04 $ & $-25.58 $ & $-25.64 $\\
CaptureTheFlag & $0.00 $ & $0.00 $ & $0.00 $ & $0.00 $ & $0.00 $ & $0.00 $\\
Earth Observation & $-14.26 $ & $-15.59 $ & $-8.75 $ & $-8.65 $ & $-7.97 $ & $-7.95 $\\
Game of Life & $483.01 $ & $484.34 $ & $522.21 $ & $527.96 $ & $544.08 $ & $566.44 $\\
KillTheKing & $-0.19 $ & $-0.16 $ & $-0.01 $ & $0.01 $ & $0.08 $ & $0.13 $\\
Manufacturer & $-1901.80 $ & $-1564.31 $ & $-1836.22 $ & $-1325.67 $ & $-1802.45 $ & $-1107.08 $\\
Navigation & $-29.78 $ & $-31.05 $ & $-23.38 $ & $-24.14 $ & $-22.19 $ & $-21.96 $\\
NumbersRace & $-0.46 $ & $-0.23 $ & $-0.28 $ & $0.34 $ & $-0.62 $ & $0.96 $\\
Othello & $-0.67 $ & $-0.50 $ & $-0.23 $ & $0.19 $ & $0.12 $ & $0.80 $\\
Pusher & $-0.00 $ & $-0.00 $ & $0.00 $ & $0.00 $ & $0.00 $ & $0.00 $\\
Push Your Luck & $53.91 $ & $53.91 $ & $54.94 $ & $54.69 $ & $53.51 $ & $53.98 $\\
Pylos & $-0.72 $ & $-0.53 $ & $-0.25 $ & $0.19 $ & $0.17 $ & $0.65 $\\
Quarto & $-0.69 $ & $-0.41 $ & $-0.41 $ & $0.21 $ & $-0.06 $ & $0.47 $\\
Cooperative Recon & $5.08 $ & $4.90 $ & $4.38 $ & $6.05 $ & $2.80 $ & $11.63 $\\
Racetrack & $-13.88 $ & $-13.79 $ & $-9.69 $ & $-9.16 $ & $-8.40 $ & $-8.27 $\\
SysAdmin & $327.42 $ & $321.87 $ & $377.52 $ & $380.58 $ & $391.39 $ & $401.75 $\\
Saving & $44.42 $ & $44.31 $ & $49.80 $ & $50.18 $ & $51.70 $ & $52.25 $\\
Skills Teaching & $-127.93 $ & $-107.20 $ & $-83.43 $ & $26.34 $ & $-84.15 $ & $70.82 $\\
Sailing Wind & $-82.40 $ & $-81.17 $ & $-66.00 $ & $-65.26 $ & $-62.23 $ & $-61.87 $\\
Tamarisk & $-868.88 $ & $-855.06 $ & $-756.94 $ & $-624.15 $ & $-730.61 $ & $-520.38 $\\
Traffic & $-23.01 $ & $-23.50 $ & $-15.48 $ & $-15.91 $ & $-12.72 $ & $-12.86 $\\
Triangle Tireworld & $28.53 $ & $49.45 $ & $36.18 $ & $77.29 $ & $34.42 $ & $82.59 $\\
TicTacToe & $-0.30 $ & $-0.14 $ & $-0.24 $ & $0.01 $ & $-0.11 $ & $0.02 $\\
Wildlife Preserve & $1193.52 $ & $1177.04 $ & $1349.91 $ & $1332.29 $ & $1375.04 $ & $1376.04 $\\

\bottomrule
\end{tabular}
}
\caption{Performance comparisons between the Vanilla UCT and Global Std parameter settings with the highest pairings score in Tab.~\ref{fig:score_gen}. Global Std has an average 40\% performance increase over Vanilla UCT in this setting.}
\label{tab:performances_versus_gen}
\end{table}

\subsection{Normalized pairing score matrices for all $\lambda$-strategies}

\begin{table}[H]\centering
\caption{The normalized pairings score matrix for each parameter-optimized $\lambda$-strategy pairing when constructed over all model-iteration budget pairs. Note that Global Std (our method) outperforms all other methods.}
\label{tab:score_matrix_all}
\scalebox{0.9}{
\setlength{\tabcolsep}{1mm}\begin{tabular}{ c c c c c c c c c c c c c }
\toprule
&Global Abs & Global Range & Global Std& Layer Abs & Layer Range & Layer Std& Local Q & Local Abs & Local Range & Local Std& Poly-UCB1 & Vanilla UCT\\
\midrule
Global Abs & $0.00 $ & $-0.30 $ & $-0.31 $ & $0.14 $ & $0.13 $ & $0.33 $ & $0.81 $ & $0.49 $ & $0.19 $ & $0.07 $ & $0.42 $ & $-0.18 $\\
Global Range & $0.30 $ & $0.00 $ & $-0.24 $ & $0.38 $ & $0.25 $ & $0.31 $ & $0.81 $ & $0.65 $ & $0.26 $ & $0.10 $ & $0.49 $ & $0.00 $\\
Global Std& $0.31 $ & $0.24 $ & $0.00 $ & $0.48 $ & $0.44 $ & $0.48 $ & $0.83 $ & $0.65 $ & $0.50 $ & $0.29 $ & $0.58 $ & $0.18 $\\
Layer Abs & $-0.14 $ & $-0.38 $ & $-0.48 $ & $0.00 $ & $0.14 $ & $0.20 $ & $0.77 $ & $0.50 $ & $0.13 $ & $0.08 $ & $0.50 $ & $-0.35 $\\
Layer Range & $-0.13 $ & $-0.25 $ & $-0.44 $ & $-0.14 $ & $0.00 $ & $-0.19 $ & $0.88 $ & $0.35 $ & $-0.21 $ & $-0.52 $ & $0.54 $ & $-0.18 $\\
Layer Std& $-0.33 $ & $-0.31 $ & $-0.48 $ & $-0.20 $ & $0.19 $ & $0.00 $ & $0.89 $ & $0.30 $ & $-0.30 $ & $-0.49 $ & $0.57 $ & $-0.20 $\\
Local Q & $-0.81 $ & $-0.81 $ & $-0.83 $ & $-0.77 $ & $-0.88 $ & $-0.89 $ & $0.00 $ & $-0.73 $ & $-0.85 $ & $-0.87 $ & $-0.56 $ & $-0.80 $\\
Local Abs & $-0.49 $ & $-0.65 $ & $-0.65 $ & $-0.50 $ & $-0.35 $ & $-0.30 $ & $0.73 $ & $0.00 $ & $-0.45 $ & $-0.54 $ & $0.40 $ & $-0.55 $\\
Local Range & $-0.19 $ & $-0.26 $ & $-0.50 $ & $-0.13 $ & $0.21 $ & $0.30 $ & $0.85 $ & $0.45 $ & $0.00 $ & $-0.20 $ & $0.57 $ & $-0.17 $\\
Local Std& $-0.07 $ & $-0.10 $ & $-0.29 $ & $-0.08 $ & $0.52 $ & $0.49 $ & $0.87 $ & $0.54 $ & $0.20 $ & $0.00 $ & $0.60 $ & $-0.15 $\\
Poly-UCB1 & $-0.42 $ & $-0.49 $ & $-0.58 $ & $-0.50 $ & $-0.54 $ & $-0.57 $ & $0.56 $ & $-0.40 $ & $-0.57 $ & $-0.60 $ & $0.00 $ & $-0.50 $\\
Vanilla UCT & $0.18 $ & $0.00 $ & $-0.18 $ & $0.35 $ & $0.18 $ & $0.20 $ & $0.80 $ & $0.55 $ & $0.17 $ & $0.15 $ & $0.50 $ & $0.00 $\\
\bottomrule
\end{tabular}}
\end{table}

\begin{table}[H]\centering
\caption{The normalized pairings score matrix for each parameter-optimized $\lambda$-strategy pairing when constructed over all models using 100 MCTS iterations. Note that Global Std (our method) outperforms all other methods.}
\label{tab:score_matrix_100}
\scalebox{0.9}{
\setlength{\tabcolsep}{1mm}\begin{tabular}{ c c c c c c c c c c c c c }
\toprule
&Global Abs & Global Range & Global Std& Layer Abs & Layer Range & Layer Std& Local Q & Local Abs & Local Range & Local Std& Poly-UCB1 & Vanilla UCT\\
\midrule
Global Abs & $0.00 $ & $-0.25 $ & $-0.21 $ & $0.00 $ & $0.07 $ & $0.21 $ & $0.64 $ & $0.46 $ & $0.25 $ & $0.07 $ & $0.25 $ & $-0.11 $\\
Global Range & $0.25 $ & $0.00 $ & $-0.18 $ & $0.11 $ & $0.18 $ & $0.25 $ & $0.68 $ & $0.64 $ & $0.36 $ & $0.04 $ & $0.21 $ & $0.14 $\\
Global Std& $0.21 $ & $0.18 $ & $0.00 $ & $0.29 $ & $0.36 $ & $0.36 $ & $0.64 $ & $0.54 $ & $0.54 $ & $0.14 $ & $0.39 $ & $0.18 $\\
Layer Abs & $0.00 $ & $-0.11 $ & $-0.29 $ & $0.00 $ & $0.00 $ & $0.14 $ & $0.64 $ & $0.61 $ & $0.11 $ & $0.07 $ & $0.39 $ & $-0.18 $\\
Layer Range & $-0.07 $ & $-0.18 $ & $-0.36 $ & $0.00 $ & $0.00 $ & $-0.21 $ & $0.79 $ & $0.39 $ & $0.21 $ & $-0.36 $ & $0.25 $ & $-0.04 $\\
Layer Std& $-0.21 $ & $-0.25 $ & $-0.36 $ & $-0.14 $ & $0.21 $ & $0.00 $ & $0.79 $ & $0.25 $ & $0.04 $ & $-0.29 $ & $0.39 $ & $0.04 $\\
Local Q & $-0.64 $ & $-0.68 $ & $-0.64 $ & $-0.64 $ & $-0.79 $ & $-0.79 $ & $0.00 $ & $-0.61 $ & $-0.68 $ & $-0.79 $ & $-0.46 $ & $-0.61 $\\
Local Abs & $-0.46 $ & $-0.64 $ & $-0.54 $ & $-0.61 $ & $-0.39 $ & $-0.25 $ & $0.61 $ & $0.00 $ & $-0.29 $ & $-0.43 $ & $0.21 $ & $-0.36 $\\
Local Range & $-0.25 $ & $-0.36 $ & $-0.54 $ & $-0.11 $ & $-0.21 $ & $-0.04 $ & $0.68 $ & $0.29 $ & $0.00 $ & $-0.32 $ & $0.36 $ & $-0.14 $\\
Local Std& $-0.07 $ & $-0.04 $ & $-0.14 $ & $-0.07 $ & $0.36 $ & $0.29 $ & $0.79 $ & $0.43 $ & $0.32 $ & $0.00 $ & $0.39 $ & $0.04 $\\
Poly-UCB1 & $-0.25 $ & $-0.21 $ & $-0.39 $ & $-0.39 $ & $-0.25 $ & $-0.39 $ & $0.46 $ & $-0.21 $ & $-0.36 $ & $-0.39 $ & $0.00 $ & $-0.21 $\\
Vanilla UCT & $0.11 $ & $-0.14 $ & $-0.18 $ & $0.18 $ & $0.04 $ & $-0.04 $ & $0.61 $ & $0.36 $ & $0.14 $ & $-0.04 $ & $0.21 $ & $0.00 $\\
\bottomrule
\end{tabular}}
\end{table}

\begin{table}[H]\centering
\caption{The normalized pairings score matrix for each parameter-optimized $\lambda$-strategy pairing when constructed over all models using 500 MCTS iterations. Note that Global Std (our method) outperforms all other methods.}
\label{tab:score_matrix_500}
\scalebox{0.9}{
\setlength{\tabcolsep}{1mm}\begin{tabular}{ c c c c c c c c c c c c c }
\toprule
&Global Abs & Global Range & Global Std& Layer Abs & Layer Range & Layer Std& Local Q & Local Abs & Local Range & Local Std& Poly-UCB1 & Vanilla UCT\\
\midrule
Global Abs & $0.00 $ & $-0.25 $ & $-0.39 $ & $0.18 $ & $0.14 $ & $0.32 $ & $0.89 $ & $0.61 $ & $0.14 $ & $-0.04 $ & $0.46 $ & $-0.32 $\\
Global Range & $0.25 $ & $0.00 $ & $-0.43 $ & $0.57 $ & $0.18 $ & $0.29 $ & $0.93 $ & $0.71 $ & $0.11 $ & $0.07 $ & $0.64 $ & $-0.32 $\\
Global Std& $0.39 $ & $0.43 $ & $0.00 $ & $0.64 $ & $0.54 $ & $0.50 $ & $0.93 $ & $0.79 $ & $0.46 $ & $0.43 $ & $0.64 $ & $0.18 $\\
Layer Abs & $-0.18 $ & $-0.57 $ & $-0.64 $ & $0.00 $ & $0.32 $ & $0.21 $ & $0.86 $ & $0.50 $ & $0.18 $ & $0.00 $ & $0.57 $ & $-0.39 $\\
Layer Range & $-0.14 $ & $-0.18 $ & $-0.54 $ & $-0.32 $ & $0.00 $ & $-0.32 $ & $0.89 $ & $0.25 $ & $-0.39 $ & $-0.68 $ & $0.75 $ & $-0.25 $\\
Layer Std& $-0.32 $ & $-0.29 $ & $-0.50 $ & $-0.21 $ & $0.32 $ & $0.00 $ & $0.93 $ & $0.43 $ & $-0.32 $ & $-0.50 $ & $0.71 $ & $-0.18 $\\
Local Q & $-0.89 $ & $-0.93 $ & $-0.93 $ & $-0.86 $ & $-0.89 $ & $-0.93 $ & $0.00 $ & $-0.79 $ & $-0.96 $ & $-0.93 $ & $-0.50 $ & $-0.96 $\\
Local Abs & $-0.61 $ & $-0.71 $ & $-0.79 $ & $-0.50 $ & $-0.25 $ & $-0.43 $ & $0.79 $ & $0.00 $ & $-0.54 $ & $-0.64 $ & $0.50 $ & $-0.82 $\\
Local Range & $-0.14 $ & $-0.11 $ & $-0.46 $ & $-0.18 $ & $0.39 $ & $0.32 $ & $0.96 $ & $0.54 $ & $0.00 $ & $-0.32 $ & $0.68 $ & $-0.11 $\\
Local Std& $0.04 $ & $-0.07 $ & $-0.43 $ & $0.00 $ & $0.68 $ & $0.50 $ & $0.93 $ & $0.64 $ & $0.32 $ & $0.00 $ & $0.71 $ & $-0.25 $\\
Poly-UCB1 & $-0.46 $ & $-0.64 $ & $-0.64 $ & $-0.57 $ & $-0.75 $ & $-0.71 $ & $0.50 $ & $-0.50 $ & $-0.68 $ & $-0.71 $ & $0.00 $ & $-0.61 $\\
Vanilla UCT & $0.32 $ & $0.32 $ & $-0.18 $ & $0.39 $ & $0.25 $ & $0.18 $ & $0.96 $ & $0.82 $ & $0.11 $ & $0.25 $ & $0.61 $ & $0.00 $\\
\bottomrule
\end{tabular}}
\end{table}

\begin{table}[H]\centering
\caption{The normalized pairings score matrix for each parameter-optimized $\lambda$-strategy pairing when constructed over all models using 2500 MCTS iterations. Note that Global Std (our method) outperforms all other methods.}
\label{tab:score_matrix_2500}
\scalebox{0.9}{
\setlength{\tabcolsep}{1mm}\begin{tabular}{ c c c c c c c c c c c c c }
\toprule
&Global Abs & Global Range & Global Std& Layer Abs & Layer Range & Layer Std& Local Q & Local Abs & Local Range & Local Std& Poly-UCB1 & Vanilla UCT\\
\midrule
Global Abs & $0.00 $ & $-0.39 $ & $-0.32 $ & $0.25 $ & $0.18 $ & $0.46 $ & $0.89 $ & $0.39 $ & $0.18 $ & $0.18 $ & $0.54 $ & $-0.11 $\\
Global Range & $0.39 $ & $0.00 $ & $-0.11 $ & $0.46 $ & $0.39 $ & $0.39 $ & $0.82 $ & $0.61 $ & $0.32 $ & $0.18 $ & $0.61 $ & $0.18 $\\
Global Std& $0.32 $ & $0.11 $ & $0.00 $ & $0.50 $ & $0.43 $ & $0.57 $ & $0.93 $ & $0.64 $ & $0.50 $ & $0.29 $ & $0.71 $ & $0.18 $\\
Layer Abs & $-0.25 $ & $-0.46 $ & $-0.50 $ & $0.00 $ & $0.11 $ & $0.25 $ & $0.82 $ & $0.39 $ & $0.11 $ & $0.18 $ & $0.54 $ & $-0.46 $\\
Layer Range & $-0.18 $ & $-0.39 $ & $-0.43 $ & $-0.11 $ & $0.00 $ & $-0.04 $ & $0.96 $ & $0.39 $ & $-0.46 $ & $-0.54 $ & $0.61 $ & $-0.25 $\\
Layer Std& $-0.46 $ & $-0.39 $ & $-0.57 $ & $-0.25 $ & $0.04 $ & $0.00 $ & $0.96 $ & $0.21 $ & $-0.61 $ & $-0.68 $ & $0.61 $ & $-0.46 $\\
Local Q & $-0.89 $ & $-0.82 $ & $-0.93 $ & $-0.82 $ & $-0.96 $ & $-0.96 $ & $0.00 $ & $-0.79 $ & $-0.89 $ & $-0.89 $ & $-0.71 $ & $-0.82 $\\
Local Abs & $-0.39 $ & $-0.61 $ & $-0.64 $ & $-0.39 $ & $-0.39 $ & $-0.21 $ & $0.79 $ & $0.00 $ & $-0.54 $ & $-0.54 $ & $0.50 $ & $-0.46 $\\
Local Range & $-0.18 $ & $-0.32 $ & $-0.50 $ & $-0.11 $ & $0.46 $ & $0.61 $ & $0.89 $ & $0.54 $ & $0.00 $ & $0.04 $ & $0.68 $ & $-0.25 $\\
Local Std& $-0.18 $ & $-0.18 $ & $-0.29 $ & $-0.18 $ & $0.54 $ & $0.68 $ & $0.89 $ & $0.54 $ & $-0.04 $ & $0.00 $ & $0.68 $ & $-0.25 $\\
Poly-UCB1 & $-0.54 $ & $-0.61 $ & $-0.71 $ & $-0.54 $ & $-0.61 $ & $-0.61 $ & $0.71 $ & $-0.50 $ & $-0.68 $ & $-0.68 $ & $0.00 $ & $-0.68 $\\
Vanilla UCT & $0.11 $ & $-0.18 $ & $-0.18 $ & $0.46 $ & $0.25 $ & $0.46 $ & $0.82 $ & $0.46 $ & $0.25 $ & $0.25 $ & $0.68 $ & $0.00 $\\
\bottomrule
\end{tabular}}
\end{table}

\subsection{Parameter-optimized performances for individual iteration budgets}.

\begin{table}[H]\centering
\caption{The pairings score for each parameter-optimized $\lambda$-strategy for each MCTS iteration budget setting sorted by the performances in the 2500 iterations setting. Note that our newly proposed method Global Std performs best overall, followed by Global Range and Vanilla UCT.}
\scalebox{1.0}{
\setlength{\tabcolsep}{1mm}\begin{tabular}{ c c c c }
\toprule
$\lambda$-strategy &100 iterations & 500 iterations & 2500 iterations\\
\midrule
Global Std& \textbf{0.35} & \textbf{0.54} & \textbf{0.47} \\
Global Range & $0.24 $ & $0.27 $ & $0.39 $\\
Vanilla UCT & $0.11 $ & $0.37 $ & $0.31 $\\
Global Abs & $0.13 $ & $0.16 $ & $0.20 $\\
Local Std& $0.21 $ & $0.28 $ & $0.20 $\\
Local Range & $-0.06 $ & $0.14 $ & $0.17 $\\
Layer Abs & $0.13 $ & $0.08 $ & $0.06 $\\
Layer Range & $0.04 $ & $-0.08 $ & $-0.04 $\\
Layer Std& $0.04 $ & $0.01 $ & $-0.15 $\\
Local Abs & $-0.29 $ & $-0.36 $ & $-0.26 $\\
Poly-UCB1 & $-0.24 $ & $-0.53 $ & $-0.49 $\\
Local Q & $-0.67 $ & $-0.87 $ & $-0.86 $\\
\bottomrule
\end{tabular}}
\label{tab:optimized_by_iteration}
\end{table}

\subsection{Normalized pairings score for all $\lambda$-strategy exploration factor pairs}

\begin{table}[H]
  \centering
  \caption{
The normalized pairings score for each $\lambda$-strategy exploration constant C pair for each MCTS iteration budget setting as well as the all-iterations settings sorted by the performances in the all-iterations setting. Note that our newly proposed method Global Std performs best overall in the 500, 2500, and  all-iterations setting and only slightly worse than Global Range for 100 iterations.}
\label{tab:scores_strategy_and_c}
\begin{minipage}[t]{0.48\textwidth}
  \centering
  \scalebox{1.0}{
  \begin{tabular}{ c c c c c }
    \toprule
($\lambda$-strategy,$C$) &100 & 500 & 2500  & All-iterations\\
\midrule
(Global Std,2) & $0.53 $ & $\boldsymbol { 0.65 }$ & $\boldsymbol { 0.74 }$ & $\boldsymbol { 0.64 }$\\
(Global Std,1) & $0.53 $ & $0.58 $ & $0.68 $ & $0.60 $\\
(Global Range,0.5) & $\boldsymbol { 0.55 }$ & $0.61 $ & $0.61 $ & $0.59 $\\
(Global Range,0.25) & $0.53 $ & $0.56 $ & $0.58 $ & $0.56 $\\
(Global Abs,1) & $0.42 $ & $0.49 $ & $0.55 $ & $0.49 $\\
(Global Std,4) & $0.41 $ & $0.48 $ & $0.54 $ & $0.48 $\\
(Global Range,1) & $0.39 $ & $0.50 $ & $0.42 $ & $0.44 $\\
(Global Abs,2) & $0.38 $ & $0.43 $ & $0.49 $ & $0.43 $\\
(Layer Abs,1) & $0.41 $ & $0.41 $ & $0.47 $ & $0.43 $\\
(Local Std,8) & $0.36 $ & $0.48 $ & $0.44 $ & $0.43 $\\
(Local Range,2) & $0.37 $ & $0.48 $ & $0.43 $ & $0.43 $\\
(Local Std,4) & $0.40 $ & $0.44 $ & $0.39 $ & $0.41 $\\
(Global Abs,0.5) & $0.40 $ & $0.41 $ & $0.40 $ & $0.40 $\\
(Layer Std,4) & $0.41 $ & $0.40 $ & $0.39 $ & $0.40 $\\
(Layer Std,8) & $0.38 $ & $0.41 $ & $0.39 $ & $0.39 $\\
(Local Abs,0.5) & $0.40 $ & $0.37 $ & $0.33 $ & $0.37 $\\
(Local Range,4) & $0.26 $ & $0.40 $ & $0.44 $ & $0.37 $\\
(Layer Abs,2) & $0.31 $ & $0.36 $ & $0.40 $ & $0.36 $\\
(Layer Abs,0.5) & $0.35 $ & $0.33 $ & $0.38 $ & $0.35 $\\
(Local Range,1) & $0.41 $ & $0.35 $ & $0.30 $ & $0.35 $\\
(Local Std,16) & $0.16 $ & $0.42 $ & $0.46 $ & $0.35 $\\
(Layer Range,2) & $0.33 $ & $0.37 $ & $0.29 $ & $0.33 $\\
(Layer Std,2) & $0.42 $ & $0.32 $ & $0.21 $ & $0.32 $\\
(Local Std,2) & $0.42 $ & $0.30 $ & $0.19 $ & $0.30 $\\
(Layer Std,16) & $0.19 $ & $0.34 $ & $0.37 $ & $0.30 $\\
(Layer Abs,0.25) & $0.38 $ & $0.27 $ & $0.24 $ & $0.30 $\\
(Layer Range,1) & $0.36 $ & $0.27 $ & $0.26 $ & $0.30 $\\
(Local Abs,1) & $0.26 $ & $0.25 $ & $0.37 $ & $0.29 $\\
(Layer Range,4) & $0.27 $ & $0.31 $ & $0.30 $ & $0.29 $\\
(Layer Abs,4) & $0.18 $ & $0.31 $ & $0.36 $ & $0.28 $\\
(Local Abs,0.25) & $0.36 $ & $0.21 $ & $0.20 $ & $0.26 $\\
(Local Range,8) & $0.08 $ & $0.28 $ & $0.40 $ & $0.25 $\\
(Global Abs,4) & $0.14 $ & $0.31 $ & $0.31 $ & $0.25 $\\
(Global Range,0.125) & $0.28 $ & $0.22 $ & $0.25 $ & $0.25 $\\
(Global Std,8) & $0.17 $ & $0.33 $ & $0.25 $ & $0.25 $\\
(Global Abs,0.25) & $0.27 $ & $0.24 $ & $0.21 $ & $0.24 $\\
(Layer Range,8) & $0.15 $ & $0.24 $ & $0.28 $ & $0.22 $\\
(Global Range,2) & $0.20 $ & $0.23 $ & $0.21 $ & $0.21 $\\
(Global Std,0.5) & $0.32 $ & $0.15 $ & $0.15 $ & $0.21 $\\
(Local Std,32) & $0.00 $ & $0.28 $ & $0.30 $ & $0.20 $\\
(Vanilla UCT,32) & $0.16 $ & $0.24 $ & $0.05 $ & $0.15 $\\
(Vanilla UCT,16) & $0.27 $ & $0.14 $ & $0.02 $ & $0.14 $\\
(Layer Std,32) & $-0.01 $ & $0.16 $ & $0.28 $ & $0.14 $\\
(Layer Range,16) & $0.00 $ & $0.21 $ & $0.15 $ & $0.12 $\\
(Vanilla UCT,64) & $0.09 $ & $0.24 $ & $0.01 $ & $0.12 $\\
(Layer Range,0.5) & $0.11 $ & $0.18 $ & $0.05 $ & $0.11 $\\
(Layer Abs,8) & $0.04 $ & $0.17 $ & $0.09 $ & $0.10 $\\
(Local Range,16) & $-0.03 $ & $0.15 $ & $0.18 $ & $0.10 $\\
(Global Std,16) & $0.05 $ & $0.11 $ & $0.13 $ & $0.09 $\\
(Global Abs,8) & $-0.04 $ & $0.10 $ & $0.19 $ & $0.09 $\\
(Global Range,4) & $-0.02 $ & $0.07 $ & $0.15 $ & $0.07 $\\
(Vanilla UCT,8) & $0.19 $ & $0.03 $ & $-0.04 $ & $0.06 $\\
(Layer Std,64) & $-0.11 $ & $0.15 $ & $0.14 $ & $0.06 $\\
(Local Range,0.5) & $0.06 $ & $0.10 $ & $-0.01 $ & $0.05 $\\
(Layer Abs,16) & $-0.01 $ & $0.04 $ & $0.10 $ & $0.04 $\\
(Vanilla UCT,1) & $0.09 $ & $0.02 $ & $0.01 $ & $0.04 $\\
(Local Abs,2) & $-0.16 $ & $0.01 $ & $0.25 $ & $0.04 $\\
(Vanilla UCT,2) & $0.12 $ & $-0.01 $ & $-0.02 $ & $0.03 $\\
(Local Std,64) & $-0.12 $ & $0.10 $ & $0.09 $ & $0.02 $\\
(Vanilla UCT,256) & $-0.12 $ & $0.07 $ & $0.11 $ & $0.02 $\\
(Global Abs,16) & $-0.09 $ & $0.04 $ & $0.11 $ & $0.02 $\\

\bottomrule
  \end{tabular}
  }
\end{minipage}
\hfill
\begin{minipage}[t]{0.48\textwidth}
  \centering
  \scalebox{1.0}{
  \begin{tabular}{ c c c c c }
    \toprule
($\lambda$-strategy,$C$) &100 & 500 & 2500  & All-iterations\\
\midrule
(Vanilla UCT,4) & $0.21 $ & $-0.03 $ & $-0.12 $ & $0.02 $\\
(Layer Abs,0.125) & $0.17 $ & $0.01 $ & $-0.15 $ & $0.01 $\\
(Global Abs,0.125) & $0.11 $ & $-0.01 $ & $-0.15 $ & $-0.02 $\\
(Layer Std,1) & $-0.00 $ & $-0.02 $ & $-0.04 $ & $-0.02 $\\
(Local Abs,0.125) & $0.09 $ & $-0.04 $ & $-0.13 $ & $-0.03 $\\
(Global Range,8) & $-0.10 $ & $0.01 $ & $-0.01 $ & $-0.03 $\\
(Layer Range,32) & $-0.13 $ & $0.02 $ & $-0.01 $ & $-0.04 $\\
(Global Abs,32) & $-0.14 $ & $-0.01 $ & $0.02 $ & $-0.04 $\\
(Layer Abs,32) & $-0.17 $ & $-0.01 $ & $0.04 $ & $-0.05 $\\
(Global Range,16) & $-0.08 $ & $-0.04 $ & $-0.02 $ & $-0.05 $\\
(Local Range,32) & $-0.25 $ & $0.06 $ & $0.04 $ & $-0.05 $\\
(Local Std,1) & $0.00 $ & $-0.07 $ & $-0.12 $ & $-0.06 $\\
(Local Range,64) & $-0.21 $ & $-0.00 $ & $0.01 $ & $-0.07 $\\
    (Global Std,32) & $-0.12 $ & $-0.07 $ & $-0.02 $ & $-0.07 $\\
(Layer Abs,64) & $-0.16 $ & $-0.06 $ & $-0.02 $ & $-0.08 $\\
(Vanilla UCT,0.5) & $-0.03 $ & $-0.10 $ & $-0.11 $ & $-0.08 $\\
(Global Std,0.25) & $0.03 $ & $-0.14 $ & $-0.15 $ & $-0.09 $\\
(Poly-UCB1,4) & $0.07 $ & $-0.12 $ & $-0.22 $ & $-0.09 $\\
(Vanilla UCT,1000) & $-0.17 $ & $-0.08 $ & $-0.03 $ & $-0.09 $\\
(Layer Range,64) & $-0.22 $ & $-0.10 $ & $-0.01 $ & $-0.11 $\\
(Global Std,64) & $-0.14 $ & $-0.15 $ & $-0.08 $ & $-0.12 $\\
(Global Range,32) & $-0.21 $ & $-0.12 $ & $-0.07 $ & $-0.13 $\\
(Global Range,64) & $-0.25 $ & $-0.10 $ & $-0.08 $ & $-0.14 $\\
(Poly-UCB1,8) & $-0.26 $ & $-0.12 $ & $-0.06 $ & $-0.14 $\\
(Global Abs,64) & $-0.21 $ & $-0.13 $ & $-0.12 $ & $-0.15 $\\
(Layer Range,0.25) & $-0.12 $ & $-0.21 $ & $-0.23 $ & $-0.19 $\\
(Poly-UCB1,2) & $0.01 $ & $-0.31 $ & $-0.30 $ & $-0.20 $\\
(Local Range,0.25) & $-0.11 $ & $-0.23 $ & $-0.26 $ & $-0.20 $\\
(Vanilla UCT,0.25) & $-0.16 $ & $-0.21 $ & $-0.27 $ & $-0.21 $\\
(Layer Std,0.5) & $-0.12 $ & $-0.29 $ & $-0.33 $ & $-0.25 $\\
(Local Abs,4) & $-0.37 $ & $-0.31 $ & $-0.08 $ & $-0.26 $\\
(Global Std,0.125) & $-0.15 $ & $-0.29 $ & $-0.36 $ & $-0.26 $\\
(Poly-UCB1,16) & $-0.34 $ & $-0.37 $ & $-0.10 $ & $-0.27 $\\
(Poly-UCB1,1) & $-0.07 $ & $-0.38 $ & $-0.43 $ & $-0.29 $\\
(Local Std,0.5) & $-0.18 $ & $-0.37 $ & $-0.39 $ & $-0.31 $\\
(Layer Range,0.125) & $-0.22 $ & $-0.36 $ & $-0.39 $ & $-0.32 $\\
(Vanilla UCT,0.125) & $-0.24 $ & $-0.33 $ & $-0.39 $ & $-0.32 $\\
(Local Range,0.125) & $-0.21 $ & $-0.41 $ & $-0.46 $ & $-0.36 $\\
(Poly-UCB1,0.5) & $-0.17 $ & $-0.44 $ & $-0.51 $ & $-0.37 $\\
(Local Abs,8) & $-0.46 $ & $-0.42 $ & $-0.24 $ & $-0.37 $\\
(Layer Std,0.25) & $-0.24 $ & $-0.39 $ & $-0.50 $ & $-0.37 $\\
(Poly-UCB1,32) & $-0.40 $ & $-0.47 $ & $-0.26 $ & $-0.38 $\\
(Local Abs,16) & $-0.45 $ & $-0.46 $ & $-0.33 $ & $-0.41 $\\
(Poly-UCB1,0.25) & $-0.24 $ & $-0.46 $ & $-0.53 $ & $-0.41 $\\
(Local Std,0.25) & $-0.29 $ & $-0.40 $ & $-0.55 $ & $-0.41 $\\
(Local Abs,64) & $-0.46 $ & $-0.47 $ & $-0.35 $ & $-0.43 $\\
(Local Abs,32) & $-0.43 $ & $-0.48 $ & $-0.39 $ & $-0.43 $\\
(Layer Std,0.125) & $-0.29 $ & $-0.48 $ & $-0.57 $ & $-0.45 $\\
(Local Std,0.125) & $-0.31 $ & $-0.46 $ & $-0.58 $ & $-0.45 $\\
(Poly-UCB1,64) & $-0.44 $ & $-0.53 $ & $-0.40 $ & $-0.46 $\\
(Local Q,0.25) & $-0.41 $ & $-0.50 $ & $-0.50 $ & $-0.47 $\\
(Poly-UCB1,0.125) & $-0.29 $ & $-0.53 $ & $-0.59 $ & $-0.47 $\\
(Local Q,0.5) & $-0.41 $ & $-0.53 $ & $-0.48 $ & $-0.47 $\\
(Local Q,1) & $-0.42 $ & $-0.53 $ & $-0.52 $ & $-0.49 $\\
(Local Q,0.125) & $-0.42 $ & $-0.54 $ & $-0.56 $ & $-0.51 $\\
(Local Q,2) & $-0.45 $ & $-0.54 $ & $-0.57 $ & $-0.52 $\\
(Local Q,4) & $-0.47 $ & $-0.61 $ & $-0.60 $ & $-0.56 $\\
(Local Q,32) & $-0.52 $ & $-0.57 $ & $-0.61 $ & $-0.57 $\\
(Local Q,8) & $-0.55 $ & $-0.56 $ & $-0.60 $ & $-0.57 $\\
(Local Q,16) & $-0.52 $ & $-0.60 $ & $-0.61 $ & $-0.58 $\\
(Local Q,64) & $-0.52 $ & $-0.63 $ & $-0.63 $ & $-0.59 $\\
\bottomrule
  \end{tabular}
  }
\end{minipage}
\end{table}

\subsection{Problem descriptions}
\label{sec:problem_list}
The environments over which the pairings score matrices in the main section were constructed are the following.
\begin{itemize}
    \item \textbf{MDPs}: Academic Advising, Cooperative Recon, Crossing Traffic, Earth Observation, Elevators, Game of Life, Manufacturer, Navigation, Push Your Luck, Racetrack, Sailing Wind, Saving, Skill Teaching, SysAdmin, Tamarisk, Traffic, Triangle Tireworld, Wildlife Preserve.
    \item \textbf{Two-player SGs}: Capture the Flag, Chess, Connect 4, Constrictor, Kill the King, Numbers Race, Othello, Pusher, Pylos, Quarto, Tic Tac Toe.
\end{itemize}
Some of these environments can be parametrized (e.g., choosing a concrete race map for Racetrack). The concrete parameter settings can be found in the \textit{ExperimentConfigs} folder in our publicly available GitHub repository accessible at \url{https://github.com/codebro634/DynamicExplorationFactorUCT}. Most of the environments are already well-described in a survey paper by Schmöcker et al. \cite{mysurvey}. In the following, we give a description for those not contained in this survey. Each environment is either a problem from the International Probabilistic Planning Competition (IPPC) \cite{sanner2011ippc}, or a board game with only three exceptions. Constrictor is a game mode from Battlesnake \cite{battlesnake}, Numbers Race is a toy problem developed by us, and Pusher is a mode from the Stratega framework \cite{abs-2009-05643,DockhornGJL20}.

\begin{itemize}

    \item \textbf{Constrictor}: Constrictor is played on an $n$ times $n$ grid. Players take turns moving to any of the neighboring (4-neighborhood) grid cells that neither moves the player out of bounds nor hits any cell that has already been visited by any of the two players. The game ends when one player has nowhere left to move. 

    \item \textbf{Connect 4}: Connect 4 is played on grid with 7 columns and 6 rows. Each turn, one player places a stone of its color in one of the columns that is not yet filled with stones. The stone occupies the first cell in the chosen column that is not yet occupied.

\item \textbf{Cooperative Recon}: This domain models a robot having to prove the existence of life on a foreign planet. The robot is modeled as moving on a 2-dimensional grid which contains a number of objects of interest and a base. If the agent is at an object of interest, it can survey the object for the existence of water and life. The probability of a positive result of the latter is dependent on whether water has been detected. If life has been detected, the agent may photograph the object of interest which is the only way to gain a reward. Each detector may break on usage making it either unusable or decreasing its chance of working. The detectors can be repaired at the base.
    
    \item \textbf{Earth Observation}: This problem models a satellite orbiting earth. Formally, each state is a position on a 2-dimensional grid, representing the satellite's longitudinal position and the latitude the camera is aimed at as well as weather levels for some designated cells. At each step, the weather levels stochastically change independent of the agent's actions which are to idle, to take a photo of the current position, or increment/decrement the current cells $y$-position (i.e. shifting the camera focus). A reward is obtained if one of the designated cells is photographed with an amount depending on the cell's current weather condition.

    \item \textbf{Elevators}: By Elevators we refer to the Elevators-Management domain scriped in \cite{mysurvey}.

    \item \textbf{Manufacturer}: In this domain, the agent manages a manufacturing company. The agent's ultimate goal is to sell goods to customers. However, to sell a good, the agent has to first produce the good, which may require building factories and acquiring the necessary goods required for production. Additional difficulty comes from the fact that the goods' price levels vary stochastically.

    \item \textbf{Navigation}: In Navigation, the goal is to move a robot on an $n \times m$ grid from $(n,1)$ to $(n,m)$ in the least number of steps. The robot may move to any of the four adjacent tiles, however, each tile is assigned a unique probability with which the robot is reset back to $(n,1)$.

    \item \textbf{Numbers Race}: In Numbers Race, players take turns choosing an integer between 1 and $n \in \mathbb{N}$. The goal is to choose a number $m \leq n$ such that the sum of all previous numbers is equal to some goal number $g \in \mathbb{N}$. If this sum exceeds $g$, then the player that overshot, loses.

\item \textbf{Push Your Luck}: In Push Your Luck the agent has to decide which of $n$, $m$-sided, not-necessarily fair dice or cash-out. If cashed-out, the agent receives a reward dependent on all dice faces that are marked. Faces are marked if they have been rolled (each face is shared by all $n$ dice). However, if the agent rolls an already marked face, or rolls two unmarked faces at the same time, all markings are removed.

\item \textbf{Quarto}: Quarto is played on a 4x4 grid. The goal is to complete a vertical, horizontal, or diagonal line of length 4 with stones that all share a common property. There are initially 16 stones that can be placed on the board. The set of stones is given by $S = \{0,1\}^4$ where the $i$-th component of a stone is referred to as a property. Players take turns placing a stone and then choosing one of the remaining, not yet placed stones which the opponent must place in the next turn (the game starts with one player selecting a stone for the opponent).
        
\item \textbf{Pylos}: Pylos is played on an initially empty 4x4x4-dimensional grid on which players take turns placing stones of their own color. Each player starts with 15 stones, which can be placed at any empty cell that is either at the bottom or contains exactly 4 stones in the layer beneath it (i.e., the stone requires a foundation). Instead of placing one of one's remaining stones, one may also move a stone of one's color any number of layers upward as long as they are not part of the foundation for another stone. The game ends when one player runs out of stones or when one player completes the pyramid (bottom layer is 4x4, then 3x3, $\dots$, 1x1). 

\item \textbf{Red Finned Blue Eye}: In this environment, the agent is tasked with preserving and restoring the Red Finned Blue Eye (RFBE) fish population which is being threatened by an invasive species of Gambusian fish. The ecosystem is being modelled as springs that are connected in a directed graph, however, the connections' accessibility is dependent on the current global water level which changes stochastically. Gambusian spreads aggressively between connected springs. 

\item \textbf{Skill Teaching}
In Skill Teaching, the agent takes the role of a tutor that is tasked with increasing the proficiency level of a student at various skills. The student can have one of three proficiency levels at each skill: Low, medium, and high. The skills from a prerequisite graph, giving the student higher chances of learning a new skill the higher the prerequisites' levels of proficiency. Difficulty arises from the proficiency levels decaying if the corresponding skill wasn't practised. This decay is deterministic for skills at medium proficiency and stochastic for those at high proficiency.

 \item \textbf{Pusher}: In Pusher, one controls several units with the goal of pushing the opponent's units into holes that are spread around the map.

    \item \textbf{Tic Tac Toe}: Tic Tac Toe is played on a grid of width and height 3. Each player is assigned a color and places one stone of its color in one of the empty grid cells. 
The first player to create a vertical, horizontal, or diagonal row of three same-colored stones wins.

\item \textbf{Traffic}:
In this environment, the agent is tasked with simultaneously controlling a number of traffic lights with the goal of minimizing traffic jams. This traffic is modelled as a directed graph, however, some edges are only available depending on the state of a traffic light. Each vertex may either contain a car or not.

    \item \textbf{Triangle Tireworld:} Triangle Tireworld refers to Tireword in \cite{mysurvey}.

    \item \textbf{Wildfire}: Wildfire models the spread of a fire on a grid. Each grid cell is either untouched, burning, or out-of-fuel meaning that no new fire can ignite at this cell. If a cell is untouched it can at each time step randomly ignite with the probability increasing exponentially in the number of neighboring burning cells. The neighborhood is defined on an instance level with most instances choosing the 8-neighborhood and manually cutting a handful of neighborhood connections between individual cells. The agent is tasked with controlling the spread of the fire.

    \item \textbf{Wildlife Preserve}: In Wildlife Preserve, the agent manages rangers to defend areas from poachers. If a ranger was sent to defend an area, and a poacher decided to attack, the poacher is caught and can not attack the area in the next step.
Each poacher has different area preferences and remembers how often which area was defended in the last couple of steps.

\end{itemize}

\subsection{Full data tables}
\label{sec:ful_data_tables}
\setcounter{page}{6}
\begin{table}[H]\centering
\caption{Part 1/3 of the full experimental data table containing the average return and 99\% confidence interval for each exploration constant-strategy-model triple with 100 MCTS iterations..}
\scalebox{0.6}{
\setlength{\tabcolsep}{1mm}\begin{tabular}{ c c c c c c c c c c }
\toprule
&Cooperative Recon & Crossing Traffic & Game of Life & Manufacturer & Navigation & Racetrack & Skills Teaching & Tamarisk & Triangle Tireworld\\
\midrule
(Global Abs,0.125) & $2.85 \pm 0.19$ & $-25.9 \pm 0.9$ & $486.1 \pm 2.2$ & $-1623.9 \pm 17.0$ & $-31.2 \pm 0.6$ & $-14.6 \pm 0.2$ & $-123.2 \pm 5.2$ & $-854.8 \pm 5.5$ & $20.5 \pm 2.4$\\
(Global Abs,0.25) & $3.13 \pm 0.20$ & $-26.4 \pm 0.9$ & $484.5 \pm 2.2$ & $-1573.0 \pm 15.5$ & $-30.2 \pm 0.6$ & $-14.1 \pm 0.2$ & $-118.3 \pm 5.3$ & $-867.4 \pm 5.6$ & $21.8 \pm 2.4$\\
(Global Abs,0.5) & $3.94 \pm 0.23$ & $-26.0 \pm 0.9$ & $482.9 \pm 2.1$ & $-1558.1 \pm 15.6$ & $-29.7 \pm 0.6$ & $-14.2 \pm 0.2$ & $-119.4 \pm 5.4$ & $-883.2 \pm 5.5$ & $34.3 \pm 2.3$\\
(Global Abs,1) & $4.96 \pm 0.25$ & $-25.5 \pm 0.9$ & $478.9 \pm 2.1$ & $-1580.9 \pm 16.5$ & $-29.7 \pm 0.6$ & $-14.4 \pm 0.2$ & $-129.1 \pm 5.6$ & $-894.3 \pm 5.3$ & $49.2 \pm 2.1$\\
(Global Abs,16) & $4.91 \pm 0.24$ & $-25.1 \pm 0.9$ & $473.6 \pm 2.2$ & $-1610.2 \pm 17.1$ & $-30.8 \pm 0.6$ & $-15.0 \pm 0.2$ & $-136.3 \pm 5.6$ & $-904.9 \pm 5.2$ & $40.6 \pm 2.2$\\
(Global Abs,2) & $5.28 \pm 0.26$ & $-25.2 \pm 0.9$ & $476.3 \pm 2.2$ & $-1585.1 \pm 16.4$ & $-30.3 \pm 0.6$ & $-14.7 \pm 0.2$ & $-127.0 \pm 5.6$ & $-902.3 \pm 5.3$ & $49.1 \pm 2.1$\\
(Global Abs,32) & $5.25 \pm 0.24$ & $-25.8 \pm 0.9$ & $475.3 \pm 2.2$ & $-1593.5 \pm 16.7$ & $-30.3 \pm 0.6$ & $-14.9 \pm 0.2$ & $-132.4 \pm 5.4$ & $-906.7 \pm 5.1$ & $40.6 \pm 2.2$\\
\bottomrule
\end{tabular}}
\label{tab:best_performances1_100}
\end{table}

\begin{table}[H]\centering
\caption{Part 2/3 of the full experimental data table containing the average return and 99\% confidence interval for each exploration constant-strategy-model triple with 100 MCTS iterations..}
\scalebox{0.6}{
\setlength{\tabcolsep}{1mm}\begin{tabular}{ c c c c c c c c c }
\toprule
&Academic Advising & Earth Observation & Push Your Luck & Saving & SysAdmin & TicTacToe & Traffic & Wildlife Preserve\\
\midrule
(Global Abs,0.125) & $-107.0 \pm 1.0$ & $-15.4 \pm 0.3$ & $52.7 \pm 0.4$ & $36.3 \pm 0.2$ & $325.8 \pm 1.9$ & $-0.30 \pm 0.01$ & $-22.1 \pm 0.3$ & $1176.3 \pm 3.0$\\
(Global Abs,0.25) & $-107.0 \pm 1.0$ & $-13.8 \pm 0.3$ & $53.4 \pm 0.4$ & $37.8 \pm 0.2$ & $327.0 \pm 2.0$ & $-0.20 \pm 0.01$ & $-22.5 \pm 0.3$ & $1163.5 \pm 2.8$\\
(Global Abs,0.5) & $-109.4 \pm 1.1$ & $-13.9 \pm 0.3$ & $54.2 \pm 0.4$ & $40.6 \pm 0.2$ & $320.8 \pm 2.0$ & $-0.14 \pm 0.01$ & $-23.4 \pm 0.3$ & $1158.3 \pm 2.7$\\
(Global Abs,1) & $-112.9 \pm 1.1$ & $-14.1 \pm 0.2$ & $53.8 \pm 0.4$ & $43.7 \pm 0.2$ & $313.3 \pm 2.0$ & $-0.12 \pm 0.01$ & $-25.1 \pm 0.3$ & $1150.9 \pm 2.6$\\
(Global Abs,16) & $-117.2 \pm 1.1$ & $-14.5 \pm 0.3$ & $52.5 \pm 0.5$ & $44.0 \pm 0.2$ & $305.9 \pm 2.0$ & $-0.30 \pm 0.01$ & $-26.1 \pm 0.3$ & $1148.7 \pm 2.6$\\
(Global Abs,2) & $-116.3 \pm 1.2$ & $-14.5 \pm 0.3$ & $53.2 \pm 0.4$ & $44.4 \pm 0.2$ & $308.8 \pm 2.0$ & $-0.16 \pm 0.01$ & $-25.7 \pm 0.3$ & $1153.0 \pm 2.6$\\
(Global Abs,32) & $-117.7 \pm 1.1$ & $-14.5 \pm 0.3$ & $52.1 \pm 0.5$ & $44.1 \pm 0.2$ & $305.7 \pm 2.0$ & $-0.31 \pm 0.01$ & $-26.3 \pm 0.3$ & $1148.7 \pm 2.6$\\
\bottomrule
\end{tabular}}
\label{tab:best_performances2_100}
\end{table}

\begin{table}[H]\centering
\caption{Part 3/3 of the full experimental data table containing the average return and 99\% confidence interval for each exploration constant-strategy-model triple with 100 MCTS iterations..}
\scalebox{0.6}{
\setlength{\tabcolsep}{1mm}\begin{tabular}{ c c c c c c c c c c c }
\toprule
&CaptureTheFlag & Chess & Connect4 & Constrictor & KillTheKing & NumbersRace & Othello & Pusher & Pylos & Quarto\\
\midrule
(Global Abs,0.125) & $\boldsymbol { 0.00 \pm 0.00 }$ & $-0.37 \pm 0.02$ & $-0.88 \pm 0.01$ & $-0.75 \pm 0.02$ & $-0.19 \pm 0.01$ & $-0.09 \pm 0.03$ & $-0.69 \pm 0.02$ & $-0.00 \pm 0.00$ & $-0.80 \pm 0.02$ & $-0.47 \pm 0.02$\\
(Global Abs,0.25) & $\boldsymbol { 0.00 \pm 0.00 }$ & $-0.38 \pm 0.02$ & $-0.81 \pm 0.01$ & $-0.71 \pm 0.02$ & $-0.20 \pm 0.01$ & $-0.10 \pm 0.03$ & $-0.64 \pm 0.02$ & $-0.00 \pm 0.00$ & $-0.73 \pm 0.02$ & $-0.39 \pm 0.02$\\
(Global Abs,0.5) & $\boldsymbol { 0.00 \pm 0.00 }$ & $-0.42 \pm 0.02$ & $-0.69 \pm 0.02$ & $-0.58 \pm 0.02$ & $-0.18 \pm 0.01$ & $-0.13 \pm 0.03$ & $-0.52 \pm 0.02$ & $-0.00 \pm 0.00$ & $-0.62 \pm 0.02$ & $-0.31 \pm 0.02$\\
(Global Abs,1) & $\boldsymbol { 0.00 \pm 0.00 }$ & $-0.46 \pm 0.02$ & $-0.59 \pm 0.02$ & $-0.44 \pm 0.02$ & $-0.16 \pm 0.01$ & $-0.25 \pm 0.03$ & $-0.44 \pm 0.02$ & $-0.00 \pm 0.00$ & $-0.48 \pm 0.02$ & $-0.33 \pm 0.02$\\
(Global Abs,16) & $\boldsymbol { 0.00 \pm 0.00 }$ & $-0.56 \pm 0.02$ & $-0.83 \pm 0.01$ & $-0.45 \pm 0.02$ & $-0.20 \pm 0.01$ & $-0.46 \pm 0.02$ & $-0.69 \pm 0.02$ & $\boldsymbol { 0.00 \pm 0.00 }$ & $-0.69 \pm 0.02$ & $-0.68 \pm 0.02$\\
(Global Abs,2) & $\boldsymbol { 0.00 \pm 0.00 }$ & $-0.46 \pm 0.02$ & $-0.61 \pm 0.02$ & $\boldsymbol { -0.37 \pm 0.02 }$ & $-0.16 \pm 0.01$ & $-0.25 \pm 0.03$ & $-0.50 \pm 0.02$ & $-0.00 \pm 0.00$ & $-0.52 \pm 0.02$ & $-0.40 \pm 0.02$\\
(Global Abs,32) & $\boldsymbol { 0.00 \pm 0.00 }$ & $-0.56 \pm 0.02$ & $-0.83 \pm 0.01$ & $-0.47 \pm 0.02$ & $-0.19 \pm 0.01$ & $-0.51 \pm 0.02$ & $-0.67 \pm 0.02$ & $-0.00 \pm 0.00$ & $-0.71 \pm 0.02$ & $-0.70 \pm 0.02$\\
\bottomrule
\end{tabular}}
\label{tab:best_performances3_100}
\end{table}

\begin{table}[H]\centering
\caption{Part 1/3 of the full experimental data table containing the average return and 99\% confidence interval for each exploration constant-strategy-model triple with 500 MCTS iterations..}
\scalebox{0.6}{
\setlength{\tabcolsep}{1mm}\begin{tabular}{ c c c c c c c c c c }
\toprule
&Cooperative Recon & Crossing Traffic & Game of Life & Manufacturer & Navigation & Racetrack & Skills Teaching & Tamarisk & Triangle Tireworld\\
\midrule
(Global Abs,0.125) & $3.26 \pm 0.20$ & $-25.5 \pm 0.9$ & $528.2 \pm 1.9$ & $-1420.5 \pm 11.9$ & $-25.3 \pm 0.5$ & $-10.7 \pm 0.1$ & $-55.2 \pm 5.1$ & $-630.5 \pm 6.0$ & $21.5 \pm 2.4$\\
(Global Abs,0.25) & $4.07 \pm 0.22$ & $-26.7 \pm 0.9$ & $531.0 \pm 1.7$ & $-1344.6 \pm 10.7$ & $-24.3 \pm 0.5$ & $-9.96 \pm 0.07$ & $-20.5 \pm 5.3$ & $-632.2 \pm 6.1$ & $26.6 \pm 2.3$\\
(Global Abs,0.5) & $5.26 \pm 0.25$ & $-25.8 \pm 0.9$ & $526.2 \pm 1.7$ & $-1347.5 \pm 11.0$ & $-23.8 \pm 0.5$ & $-9.72 \pm 0.07$ & $5.20 \pm 5.42$ & $-646.4 \pm 6.1$ & $46.8 \pm 2.1$\\
(Global Abs,1) & $5.92 \pm 0.28$ & $-25.5 \pm 0.9$ & $522.3 \pm 1.8$ & $-1350.3 \pm 11.4$ & $-23.2 \pm 0.5$ & $-9.82 \pm 0.08$ & $2.83 \pm 5.65$ & $-654.3 \pm 6.2$ & $72.2 \pm 1.3$\\
(Global Abs,16) & $4.40 \pm 0.27$ & $-25.9 \pm 0.9$ & $518.1 \pm 1.8$ & $-1370.2 \pm 12.3$ & $-23.7 \pm 0.5$ & $-10.2 \pm 0.1$ & $-33.3 \pm 5.4$ & $-674.0 \pm 6.1$ & $70.0 \pm 1.5$\\
(Global Abs,2) & $5.35 \pm 0.29$ & $-25.5 \pm 0.9$ & $520.0 \pm 1.8$ & $-1371.6 \pm 12.1$ & $-23.7 \pm 0.5$ & $-9.94 \pm 0.08$ & $-10.2 \pm 5.5$ & $-672.6 \pm 6.1$ & $76.5 \pm 1.1$\\
(Global Abs,32) & $4.57 \pm 0.27$ & $-26.2 \pm 0.9$ & $517.2 \pm 1.8$ & $-1379.0 \pm 12.2$ & $-23.8 \pm 0.5$ & $-10.1 \pm 0.1$ & $-31.2 \pm 5.5$ & $-671.7 \pm 6.1$ & $68.2 \pm 1.5$\\
\bottomrule
\end{tabular}}
\label{tab:best_performances1_500}
\end{table}

\begin{table}[H]\centering
\caption{Part 2/3 of the full experimental data table containing the average return and 99\% confidence interval for each exploration constant-strategy-model triple with 500 MCTS iterations..}
\scalebox{0.6}{
\setlength{\tabcolsep}{1mm}\begin{tabular}{ c c c c c c c c c }
\toprule
&Academic Advising & Earth Observation & Push Your Luck & Saving & SysAdmin & TicTacToe & Traffic & Wildlife Preserve\\
\midrule
(Global Abs,0.125) & $-78.7 \pm 0.7$ & $-10.0 \pm 0.2$ & $53.4 \pm 0.4$ & $37.1 \pm 0.2$ & $374.1 \pm 1.6$ & $-0.08 \pm 0.01$ & $-15.5 \pm 0.2$ & $1330.8 \pm 3.1$\\
(Global Abs,0.25) & $-74.8 \pm 0.7$ & $-8.59 \pm 0.18$ & $54.0 \pm 0.4$ & $39.5 \pm 0.2$ & $382.4 \pm 1.5$ & $-0.03 \pm 0.01$ & $-15.3 \pm 0.2$ & $1311.0 \pm 3.5$\\
(Global Abs,0.5) & $-75.2 \pm 0.7$ & $-8.67 \pm 0.19$ & $54.0 \pm 0.3$ & $44.4 \pm 0.2$ & $380.0 \pm 1.5$ & $-0.00 \pm 0.01$ & $-15.8 \pm 0.2$ & $1298.3 \pm 3.7$\\
(Global Abs,1) & $-77.6 \pm 0.7$ & $-8.72 \pm 0.19$ & $54.1 \pm 0.3$ & $49.3 \pm 0.1$ & $376.1 \pm 1.5$ & $0.01 \pm 0.01$ & $-16.6 \pm 0.2$ & $1291.9 \pm 3.8$\\
(Global Abs,16) & $-80.3 \pm 0.8$ & $-8.72 \pm 0.19$ & $55.2 \pm 0.4$ & $48.8 \pm 0.1$ & $367.3 \pm 1.6$ & $-0.23 \pm 0.01$ & $-17.7 \pm 0.3$ & $1285.2 \pm 3.8$\\
(Global Abs,2) & $-78.8 \pm 0.7$ & $-8.81 \pm 0.19$ & $54.7 \pm 0.3$ & $49.8 \pm 0.1$ & $372.7 \pm 1.6$ & $0.01 \pm 0.01$ & $-16.9 \pm 0.2$ & $1288.4 \pm 3.8$\\
(Global Abs,32) & $-80.2 \pm 0.7$ & $-8.87 \pm 0.19$ & $55.4 \pm 0.4$ & $48.7 \pm 0.1$ & $368.0 \pm 1.6$ & $-0.24 \pm 0.01$ & $-17.5 \pm 0.2$ & $1286.9 \pm 3.8$\\
\bottomrule
\end{tabular}}
\label{tab:best_performances2_500}
\end{table}

\begin{table}[H]\centering
\caption{Part 3/3 of the full experimental data table containing the average return and 99\% confidence interval for each exploration constant-strategy-model triple with 500 MCTS iterations..}
\scalebox{0.6}{
\setlength{\tabcolsep}{1mm}\begin{tabular}{ c c c c c c c c c c c }
\toprule
&CaptureTheFlag & Chess & Connect4 & Constrictor & KillTheKing & NumbersRace & Othello & Pusher & Pylos & Quarto\\
\midrule
(Global Abs,0.125) & $\boldsymbol { 0.00 \pm 0.00 }$ & $0.65 \pm 0.02$ & $-0.62 \pm 0.02$ & $-0.63 \pm 0.02$ & $-0.09 \pm 0.01$ & $0.75 \pm 0.02$ & $-0.44 \pm 0.02$ & $-0.00 \pm 0.00$ & $-0.61 \pm 0.02$ & $-0.02 \pm 0.02$\\
(Global Abs,0.25) & $\boldsymbol { 0.00 \pm 0.00 }$ & $\boldsymbol { 0.67 \pm 0.02 }$ & $-0.47 \pm 0.02$ & $-0.56 \pm 0.02$ & $-0.08 \pm 0.01$ & $0.82 \pm 0.01$ & $-0.27 \pm 0.02$ & $0.00 \pm 0.00$ & $-0.43 \pm 0.02$ & $0.08 \pm 0.02$\\
(Global Abs,0.5) & $\boldsymbol { 0.00 \pm 0.00 }$ & $0.64 \pm 0.02$ & $-0.15 \pm 0.02$ & $-0.38 \pm 0.02$ & $-0.05 \pm 0.01$ & $0.79 \pm 0.02$ & $0.08 \pm 0.03$ & $0.00 \pm 0.00$ & $-0.08 \pm 0.03$ & $0.17 \pm 0.02$\\
(Global Abs,1) & $\boldsymbol { 0.00 \pm 0.00 }$ & $0.58 \pm 0.02$ & $0.25 \pm 0.02$ & $-0.02 \pm 0.03$ & $-0.02 \pm 0.01$ & $0.75 \pm 0.02$ & $0.36 \pm 0.02$ & $-0.00 \pm 0.00$ & $0.26 \pm 0.03$ & $0.24 \pm 0.02$\\
(Global Abs,16) & $\boldsymbol { 0.00 \pm 0.00 }$ & $-0.06 \pm 0.03$ & $-0.21 \pm 0.02$ & $-0.10 \pm 0.03$ & $0.00 \pm 0.01$ & $-0.27 \pm 0.03$ & $-0.21 \pm 0.03$ & $0.00 \pm 0.00$ & $-0.22 \pm 0.03$ & $-0.35 \pm 0.02$\\
(Global Abs,2) & $\boldsymbol { 0.00 \pm 0.00 }$ & $0.35 \pm 0.02$ & $0.17 \pm 0.03$ & $\boldsymbol { 0.06 \pm 0.03 }$ & $-0.00 \pm 0.01$ & $0.35 \pm 0.02$ & $0.21 \pm 0.03$ & $0.00 \pm 0.00$ & $0.20 \pm 0.03$ & $0.21 \pm 0.02$\\
(Global Abs,32) & $\boldsymbol { 0.00 \pm 0.00 }$ & $-0.08 \pm 0.03$ & $-0.24 \pm 0.02$ & $-0.11 \pm 0.03$ & $-0.01 \pm 0.01$ & $-0.29 \pm 0.02$ & $-0.26 \pm 0.02$ & $0.00 \pm 0.00$ & $-0.24 \pm 0.03$ & $-0.40 \pm 0.02$\\
\bottomrule
\end{tabular}}
\label{tab:best_performances3_500}
\end{table}

\begin{table}[H]\centering
\caption{Part 1/3 of the full experimental data table containing the average return and 99\% confidence interval for each exploration constant-strategy-model triple with 2500 MCTS iterations..}
\scalebox{0.6}{
\setlength{\tabcolsep}{1mm}\begin{tabular}{ c c c c c c c c c c }
\toprule
&Cooperative Recon & Crossing Traffic & Game of Life & Manufacturer & Navigation & Racetrack & Skills Teaching & Tamarisk & Triangle Tireworld\\
\midrule
(Global Abs,0.125) & $3.33 \pm 0.20$ & $-25.5 \pm 0.9$ & $559.1 \pm 1.8$ & $-1297.6 \pm 8.5$ & $-23.7 \pm 0.5$ & $-9.78 \pm 0.06$ & $-51.6 \pm 5.0$ & $-554.6 \pm 5.6$ & $18.7 \pm 2.4$\\
(Global Abs,0.25) & $4.52 \pm 0.23$ & $-25.4 \pm 0.9$ & $569.0 \pm 1.5$ & $-1172.2 \pm 6.0$ & $-22.4 \pm 0.5$ & $-8.78 \pm 0.03$ & $-10.7 \pm 5.2$ & $-533.0 \pm 5.7$ & $26.8 \pm 2.3$\\
(Global Abs,0.5) & $8.45 \pm 0.26$ & $-26.2 \pm 0.9$ & $564.3 \pm 1.5$ & $-1146.3 \pm 6.5$ & $-22.2 \pm 0.5$ & $-8.43 \pm 0.03$ & $28.8 \pm 5.4$ & $-526.1 \pm 5.8$ & $49.4 \pm 2.0$\\
(Global Abs,1) & $12.1 \pm 0.3$ & $-25.7 \pm 0.9$ & $562.1 \pm 1.5$ & $-1165.5 \pm 7.7$ & $-24.8 \pm 0.6$ & $-8.40 \pm 0.03$ & $57.7 \pm 5.8$ & $-531.7 \pm 5.8$ & $74.2 \pm 1.2$\\
(Global Abs,16) & $2.62 \pm 0.23$ & $-25.7 \pm 0.9$ & $557.4 \pm 1.5$ & $-1220.9 \pm 10.0$ & $-21.5 \pm 0.5$ & $-8.53 \pm 0.03$ & $30.6 \pm 5.4$ & $-537.7 \pm 5.9$ & $80.2 \pm 0.9$\\
(Global Abs,2) & $8.40 \pm 0.32$ & $-26.3 \pm 0.9$ & $559.5 \pm 1.5$ & $-1179.9 \pm 8.4$ & $-24.6 \pm 0.6$ & $-8.45 \pm 0.03$ & $64.4 \pm 5.5$ & $-533.0 \pm 5.7$ & $82.0 \pm 0.8$\\
(Global Abs,32) & $2.42 \pm 0.22$ & $-25.7 \pm 0.9$ & $557.2 \pm 1.6$ & $-1215.9 \pm 9.7$ & $-21.7 \pm 0.5$ & $-8.55 \pm 0.03$ & $29.3 \pm 5.4$ & $-536.7 \pm 5.8$ & $80.2 \pm 0.9$\\
\bottomrule
\end{tabular}}
\label{tab:best_performances1_2500}
\end{table}

\begin{table}[H]\centering
\caption{Part 2/3 of the full experimental data table containing the average return and 99\% confidence interval for each exploration constant-strategy-model triple with 2500 MCTS iterations..}
\scalebox{0.6}{
\setlength{\tabcolsep}{1mm}\begin{tabular}{ c c c c c c c c c }
\toprule
&Academic Advising & Earth Observation & Push Your Luck & Saving & SysAdmin & TicTacToe & Traffic & Wildlife Preserve\\
\midrule
(Global Abs,0.125) & $-70.7 \pm 0.7$ & $-8.69 \pm 0.18$ & $53.5 \pm 0.4$ & $37.6 \pm 0.2$ & $388.3 \pm 1.4$ & $0.01 \pm 0.01$ & $-13.8 \pm 0.2$ & $1376.3 \pm 0.6$\\
(Global Abs,0.25) & $-64.7 \pm 0.6$ & $-7.97 \pm 0.16$ & $55.1 \pm 0.4$ & $40.1 \pm 0.2$ & $402.5 \pm 1.3$ & $0.03 \pm 0.01$ & $-12.7 \pm 0.2$ & $1373.5 \pm 1.0$\\
(Global Abs,0.5) & $-64.2 \pm 0.6$ & $-7.79 \pm 0.15$ & $55.1 \pm 0.3$ & $46.3 \pm 0.2$ & $403.3 \pm 1.2$ & $0.05 \pm 0.01$ & $-13.1 \pm 0.2$ & $1372.4 \pm 1.1$\\
(Global Abs,1) & $-64.1 \pm 0.6$ & $-7.94 \pm 0.15$ & $54.5 \pm 0.2$ & $51.7 \pm 0.1$ & $400.6 \pm 1.3$ & $0.02 \pm 0.00$ & $-13.2 \pm 0.2$ & $1370.8 \pm 1.2$\\
(Global Abs,16) & $-64.4 \pm 0.6$ & $-7.91 \pm 0.15$ & $53.5 \pm 0.2$ & $51.0 \pm 0.0$ & $399.5 \pm 1.3$ & $0.02 \pm 0.00$ & $-13.7 \pm 0.2$ & $1370.1 \pm 1.3$\\
(Global Abs,2) & $-64.5 \pm 0.6$ & $-7.93 \pm 0.15$ & $53.7 \pm 0.2$ & $51.6 \pm 0.1$ & $399.4 \pm 1.3$ & $0.02 \pm 0.00$ & $-13.5 \pm 0.2$ & $1370.6 \pm 1.3$\\
(Global Abs,32) & $-64.6 \pm 0.6$ & $-7.74 \pm 0.14$ & $53.6 \pm 0.2$ & $51.0 \pm 0.0$ & $399.0 \pm 1.3$ & $-0.27 \pm 0.01$ & $-13.6 \pm 0.2$ & $1369.7 \pm 1.3$\\
\bottomrule
\end{tabular}}
\label{tab:best_performances2_2500}
\end{table}

\begin{table}[H]\centering
\caption{Part 3/3 of the full experimental data table containing the average return and 99\% confidence interval for each exploration constant-strategy-model triple with 2500 MCTS iterations..}
\scalebox{0.6}{
\setlength{\tabcolsep}{1mm}\begin{tabular}{ c c c c c c c c c c c }
\toprule
&CaptureTheFlag & Chess & Connect4 & Constrictor & KillTheKing & NumbersRace & Othello & Pusher & Pylos & Quarto\\
\midrule
(Global Abs,0.125) & $0.00 \pm 0.00$ & $0.96 \pm 0.01$ & $-0.41 \pm 0.02$ & $-0.51 \pm 0.02$ & $-0.02 \pm 0.02$ & $0.96 \pm 0.01$ & $-0.17 \pm 0.02$ & $0.00 \pm 0.00$ & $-0.41 \pm 0.02$ & $0.22 \pm 0.02$\\
(Global Abs,0.25) & $0.00 \pm 0.00$ & $0.97 \pm 0.01$ & $-0.22 \pm 0.02$ & $-0.39 \pm 0.02$ & $-0.02 \pm 0.02$ & $0.96 \pm 0.01$ & $0.09 \pm 0.02$ & $0.00 \pm 0.00$ & $-0.14 \pm 0.03$ & $0.25 \pm 0.02$\\
(Global Abs,0.5) & $0.00 \pm 0.00$ & $0.98 \pm 0.01$ & $0.26 \pm 0.02$ & $-0.24 \pm 0.03$ & $0.03 \pm 0.02$ & $0.96 \pm 0.01$ & $0.54 \pm 0.02$ & $0.00 \pm 0.00$ & $0.37 \pm 0.02$ & $0.32 \pm 0.02$\\
(Global Abs,1) & $0.00 \pm 0.00$ & $0.99 \pm 0.00$ & $0.73 \pm 0.02$ & $0.32 \pm 0.02$ & $0.08 \pm 0.02$ & $\boldsymbol { 0.97 \pm 0.01 }$ & $0.86 \pm 0.01$ & $0.00 \pm 0.00$ & $0.69 \pm 0.02$ & $0.45 \pm 0.02$\\
(Global Abs,16) & $0.00 \pm 0.00$ & $0.77 \pm 0.02$ & $0.20 \pm 0.02$ & $0.10 \pm 0.03$ & $0.12 \pm 0.02$ & $-0.59 \pm 0.02$ & $0.16 \pm 0.03$ & $0.00 \pm 0.00$ & $0.16 \pm 0.03$ & $-0.02 \pm 0.02$\\
(Global Abs,2) & $0.00 \pm 0.00$ & $0.99 \pm 0.00$ & $0.68 \pm 0.02$ & $0.39 \pm 0.02$ & $0.12 \pm 0.02$ & $0.97 \pm 0.01$ & $0.78 \pm 0.02$ & $0.00 \pm 0.00$ & $0.64 \pm 0.02$ & $0.48 \pm 0.02$\\
(Global Abs,32) & $0.00 \pm 0.00$ & $0.74 \pm 0.02$ & $0.16 \pm 0.03$ & $0.04 \pm 0.03$ & $0.10 \pm 0.02$ & $-0.70 \pm 0.02$ & $0.06 \pm 0.03$ & $0.00 \pm 0.00$ & $0.10 \pm 0.03$ & $-0.13 \pm 0.02$\\
\bottomrule
\end{tabular}}
\label{tab:bestwerew}
\end{table}

\end{document}